# Adaptive Automatic Model Selection for Demand Forecasting under Heterogeneous Demand Patterns

Adolfo González[a]* and Víctor Parada[a]

[a] Department of Computer Engineering and Informatics, Faculty of Engineering, University of Santiago of Chile, Santiago, Chile

* Corresponding author. Email: adolfo.gonzalez.c@usach.cl

ORCID
Adolfo González: https://orcid.org/0009-0008-4452-1285
Víctor Parada: https://orcid.org/0000-0002-8649-5694

E-MAIL
Adolfo González: adolfo.gonzalez.c@usach.cl
Víctor Parada: victor.parada@usach.cl

**Abstract**

Demand forecasting is critical for inventory planning, procurement, replenishment, production, and capacity decisions in heterogeneous supply chains. However, selecting the most appropriate model for each demand series remains challenging because performance varies across datasets, demand structures, horizons, and evaluation metrics. This study proposes the Adaptive Hybrid Selector (AHS), an automatic model-selection rule that combines observed predictive performance with structural demand information. AHS uses demand frequency and series variability to activate a hierarchical logic based on RMSSE, MAE, sMAPE, and BIAS. The proposed selector is compared with Overall Weighted Average (OWA), a benchmark-relative criterion based on sMAPE and MASE, and Equilibrium Ranking Aggregation (ERA), a comparator based on MAE, RMSE, and $R^2$ rankings. The empirical evaluation uses the Walmart, M3, M4, and M5 datasets, three training–testing partitions, 22 forecasting models, and horizons of up to 12 cycles. Selector performance is assessed ex post using Global Relative Accuracy (GRA), interpreted as an indicator of volumetric coherence between accumulated forecasted demand and observed demand. Results show that AHS provides the most robust overall behavior, especially in M5, while OWA remains competitive in more regular datasets. ERA shows lower volumetric coherence in most configurations. These findings suggest that automatic model selection should account for demand structure and be evaluated using ex post indicators of volumetric coherence.

## 1. Introduction

In increasingly dynamic production and supply environments, demand forecasting constitutes a critical input for inventory planning, procurement, replenishment, production scheduling, and capacity allocation. Forecast quality directly affects operational continuity, service levels, and financial efficiency, particularly when demand series exhibit variability, intermittency, structural changes, or unstable consumption patterns (Goltsos et al., 2022; González, 2020; J. Huber et al., 2017). From this perspective, forecast errors are not only statistical deviations but also potential sources of overstock, stockouts, capital immobilization, replenishment instability, and service deterioration (Silver et al., 1998; Theodorou et al., 2025; Ulrich et al., 2022).

Despite advances in statistical, machine learning, deep learning, and hybrid forecasting methods, selecting the most appropriate forecasting model for each demand series remains an open problem. Empirical evidence from large-scale forecasting competitions shows that no single model is universally dominant across datasets, frequencies, horizons, aggregation levels, or demand structures (Makridakis et al., 2018, 2022; Petropoulos et al., 2022). Moreover, different error metrics may induce conflicting model rankings because they capture different aspects of predictive performance and may correspond to different statistical functionals (Gneiting, 2011; Hyndman & Koehler, 2006; Kolassa, 2020). This makes model selection particularly challenging in multi-product environments, where manual SKU-level selection is costly, difficult to reproduce, and hard to scale (Garred et al., 2026; Poler & Mula, 2011).

Automatic model selection has emerged as a practical approach to this problem, enabling forecasting models to be systematically assigned across multiple demand series. However, many selection approaches rely mainly on observed predictive performance or aggregate rules, without explicitly incorporating demand structure into the selection logic. This is relevant because smooth, intermittent, erratic, and lumpy demand patterns may require different selection criteria, since demand frequency, variability, and zero-demand occurrence affect both forecast stability and operational usefulness (Petropoulos et al., 2022; Syntetos et al., 2005; Syntetos & Boylan, 2005). In supply chain contexts, this issue is especially important because the selected model determines the demand signal used for purchasing, replenishment, inventory exposure, production, and resource-allocation decisions.

Accordingly, the selection of a forecasting model should be interpreted as a decision-support layer rather than a purely statistical ranking task. Beyond pointwise forecast errors, it is relevant to evaluate whether the selected model generates an accumulated forecast volume that remains coherent with the observed demand over the evaluation horizon. This operational perspective is consistent with recent automatic model-selection research in supply chains, where criteria such as OWA are used not only to compare predictive accuracy but also to assess the practical value of automatic selection in terms of planning usefulness, implementation burden, and computational cost (Garred et al., 2026).

This study proposes and evaluates the Adaptive Hybrid Selector (AHS), an adaptive model-selection rule that combines demand-structure information with a hierarchical decision logic based on RMSSE, MAE, sMAPE, and BIAS. AHS is compared with two alternative selectors: Overall Weighted Average (OWA), a benchmark-relative criterion based on sMAPE and MASE, and Equilibrium Ranking Aggregation (ERA), a multicriteria ranking-aggregation comparator based on MAE, RMSE, and $R^2$. The empirical evaluation uses the Walmart, M3, M4, and M5 datasets, three training–testing partitions, 22 forecasting models, and evaluation horizons of up to 12 cycles.

The contribution of this study is threefold. First, it formalizes AHS as an adaptive automatic selection rule that integrates observed predictive performance and demand structure. Second, it compares AHS with OWA and ERA using a common, reproducible experimental design across heterogeneous demand datasets. Third, it evaluates the ex post consequences of model selection through Global Relative Accuracy (GRA), interpreted as an operational indicator of volumetric coherence between accumulated forecasted demand and observed demand. In doing so, the study connects automatic model selection for forecasting with planning, inventory, and replenishment decisions in heterogeneous-demand environments.

## 2. Literature Review

Inventory management constitutes a critical function for business competitiveness because it connects procurement, replenishment, product availability, and service continuity with the firm's competitive strategy. In multiproduct environments characterized by demand variability, heterogeneous turnover rates, and differences in product relevance, inventory management cannot be limited to reducing holding costs; it must also consider service levels, supply continuity, replenishment efficiency, and responsiveness to market requirements (Garred et al., 2026; González, 2020; Silver et al., 1998). In this context, demand forecasting transforms historical information into future estimates that support purchasing, resource planning, replenishment, and inventory allocation across the supply chain (Kumar et al., 2025; Mentzer & Moon, 2004; Sepúlveda-Rojas et al., 2015). Reliable forecasts improve operational and financial performance, whereas forecast errors may generate asymmetric effects: overestimation can increase inventory costs, obsolescence risk, and capital immobilization, while underestimation can lead to stockouts, lost sales, service deterioration, and replenishment instability (Kumar et al., 2025; Sousa et al., 2025; Theodorou et al., 2025; Ulrich et al., 2022).

Despite advances in statistical methods, machine learning, deep learning, and hybrid forecasting approaches, the systematic adoption of demand forecasting remains limited in many organizations. Although the field has achieved substantial methodological and technological progress (Petropoulos et al., 2022), practical implementation still faces organizational, technical, computational, and trust-related barriers (Goodwin et al., 2023). This limitation is particularly relevant for firms managing multiple product references, where manually selecting the best model for each demand series is complex, costly, and difficult to reproduce and scale. Consequently, the problem is not only to generate forecasts, but also to develop automatic and technically justified procedures for assigning forecasting models to individual demand series according to their predictive performance, structural characteristics, and operational use (Erjiang et al., 2022; Garred et al., 2026; Sepúlveda-Rojas et al., 2015).

A central conclusion in the forecasting literature is that no model is universally superior across all datasets, demand structures, horizons, and evaluation criteria. Evidence from forecasting competitions and comparative studies shows that the performance of statistical, machine learning, deep learning, and hybrid methods varies according to the temporal structure of the series, observation frequency, seasonality, variability, intermittency, and the metric used to assess error (Makridakis et al., 2018, 2022; Makridakis & Petropoulos, 2020; Petropoulos et al., 2022). This difficulty is intensified in heterogeneous business contexts, where demand may exhibit frequent, intermittent, erratic, or highly variable patterns, as well as structural changes and different levels of predictability (Hendry & Pretis, 2023; Syntetos et al., 2005; Syntetos & Boylan, 2005). Under these conditions, aggregate selection, understood as the use of a single forecasting model for all series, is restrictive because it assumes a global dominance that is rarely sustained empirically. Individual model selection, by contrast, seeks to identify the most appropriate method for each series, although its practical feasibility requires automatic, reproducible, and scalable procedures (Garred et al., 2026).

Forecasting model selection has been addressed through information criteria, out-of-sample validation, heuristic rules, expert intervention, feature-based selection, meta-learning, and multicriteria approaches. These procedures have contributed to systematizing model choice, but they may be limited when applied to heterogeneous environments involving multiple demand series and methods of different natures. Out-of-sample performance-based approaches are closer to operational forecasting practice because they evaluate candidate models on data not used during model fitting and allow comparisons under a common empirical criterion (Fildes & Petropoulos, 2015; Garred et al., 2026; Tashman, 2000). However, they require adequate separation between the training, selection, and test stages to avoid information leakage and ensure an independent estimate of predictive performance.

Recent forecasting literature increasingly treats model selection as a decision-support problem aimed at improving scalability, reproducibility, and operational usability in environments with many demand series. Since forecasting processes often need to be repeated across numerous products, automated or expert support systems are required to assist decision makers and reduce manual workload (Poler & Mula, 2011). This need is reinforced by automatic model-selection research, which shows that expert-driven model construction, feature engineering, and hyperparameter tuning are difficult to scale across diverse forecasting tasks (Abdallah et al., 2025). However, practical adoption also depends on data availability, technological capability, user expertise, and organizational conditions, since forecasting practices may be constrained by limited historical data, insufficient training, low technology adoption, and reliance on judgmental methods (Aidoo-Anderson et al., 2025). Therefore, model choice should consider not only predictive accuracy but also computational feasibility, interpretability, performance stability, and practical usefulness for inventory planning.

A key issue in automatic model selection is the choice of evaluation criteria. Different error measures may produce contradictory rankings, especially when models are compared across heterogeneous series or under multicriteria evaluation settings (Badulescu et al., 2021; Hyndman & Koehler, 2006; Makridakis et al., 2018, 2022). These discrepancies are not merely technical differences between equivalent indicators, since each metric reflects a different error-penalization logic and may support different decision objectives. Therefore, the evaluation criterion must be consistent with the forecast's purpose, the type of demand analyzed, and the operational use of the prediction (Gneiting, 2011; Kolassa, 2020). In this context, model selection may follow monometric, benchmark-relative, or multicriteria logic. Monometric criteria facilitate interpretation but may hide relevant conflicts among error dimensions; benchmark-relative criteria evaluate whether a model improves upon a common reference; and multicriteria criteria provide a more balanced view of global performance by integrating several evaluation dimensions.

Beyond metric-based evaluation, the literature highlights the importance of considering the structural characteristics of demand. Demand frequency, variability, zero-demand occurrence, level shifts, seasonality, and intermittency can substantially affect the relative performance of forecasting models (Petropoulos et al., 2022; Syntetos et al., 2005; Syntetos & Boylan, 2005). Therefore, adaptive selection approaches seek to ensure that the decision rule does not depend only on observed error but also on the demand regime. This perspective supports the development of adaptive hybrid selectors, such as AHS, that link model selection to the observed structure of the series. For frequent and relatively stable demand, a multicriteria evaluation that balances accuracy, stability, and bias may be appropriate. Conversely, for irregular, intermittent, or highly variable demand, more robust and conservative criteria may be preferable to avoid decisions based on unstable indicators or extreme observations.

The incorporation of demand structure into model selection is also relevant for inventory management. Low-variability and high-frequency series allow more stable planning, whereas intermittent or erratic series increase the risk of overstocking, stockouts, and inefficient replenishment. Consequently, the same magnitude of forecast error may have different operational implications depending on the demand regime. A moderate error in a high-turnover product may directly affect service continuity, whereas a similar error in an intermittent series may lead to unnecessary purchases or the accumulation of obsolete inventory. Thus, automatic model selection should consider not only statistical performance but also how such performance translates into procurement, replenishment, and inventory-control decisions (González, 2020; Kumar et al., 2025; Theodorou et al., 2025; Ulrich et al., 2022).

In summary, the literature identifies four persistent challenges in demand forecasting model selection: the absence of a universally dominant model, instability in rankings across evaluation criteria, the difficulty of automating individual selection in multiproduct environments, and the need to incorporate structural demand characteristics into the decision rule. Although information criteria, out-of-sample validation, feature-based selection, meta-learning, and expert intervention have advanced the field, inventory environments require more scalable, reproducible, and interpretable procedures. Within this framework, the present study aims to develop an adaptive hybrid selector, referred to as AHS, that assigns forecasting models based on both observed predictive performance and demand structure. Thus, model selection is formalized as an automated decision-support process to improve inventory planning in heterogeneous-demand environments.

## 3. Materials and Methods

This section presents the methodological design used to evaluate forecasting model-selection criteria under structural heterogeneity and horizon-dependent predictive performance. It describes the datasets, training–testing partitions, forecasting models, performance metrics, inferential methods, ex post volumetric assessment, and selection criteria, distinguishing monometric, multicriteria, and adaptive approaches.

### 3.1. Predictive Performance Evaluation Metrics

The evaluation of demand forecasting models requires consistency between the forecasting objective, the selected error metric, and the statistical functional elicited by that metric. Error measures are not interchangeable: squared-error metrics are generally associated with the conditional mean, absolute-error metrics with the conditional median, and percentage-based metrics may induce weighted functionals that differ from both (Gneiting, 2011; Kolassa, 2020). Accordingly, this study assigns each metric a specific methodological role rather than treating all indicators as equivalent expressions of a universal concept of accuracy.

MAE, RMSE, Mean Absolute Scaled Error (MASE), RMSSE, MAPE, symmetric Mean Absolute Percentage Error (sMAPE), $R^2$, and forecast bias (BIAS) are used as evaluation indicators. MAE captures absolute error magnitude, RMSE penalizes large deviations, and MASE and RMSSE provide scale-normalized measures for comparing heterogeneous demand series (Hyndman & Athanasopoulos, 2021; Hyndman & Koehler, 2006; Makridakis et al., 2022). Within the proposed selection logic, RMSSE serves as the primary scaled-error criterion, MAE supports absolute-error refinement, sMAPE provides relative-error diagnosis, and BIAS acts as a directional indicator of systematic overestimation or underestimation. MAPE and sMAPE are interpreted cautiously because percentage-based metrics may behave undesirably near zero values and may correspond to functionals different from those induced by absolute or squared errors (Gneiting, 2011; Hyndman & Koehler, 2006; Kolassa, 2020; Kolassa & Schütz, 2007). $R^2$ is retained only as a descriptive indicator of explanatory capacity, while BIAS is relevant for inventory exposure, replenishment risk, service reliability, and supply chain costs (Doszyń & Dudek, 2024; Sanders & Graman, 2016).

Thus, the metrics are organized hierarchically within the adaptive selection framework. RMSSE provides the main scale-normalized comparison, MAE supports absolute-error refinement, sMAPE refines relative-error performance, and BIAS is used for directional diagnosis or tie-breaking. This hierarchy is particularly relevant because frequent and relatively stable demand can support multicriteria refinement, whereas intermittent, irregular, or highly variable demand may require more conservative criteria to avoid unstable decisions driven by sparse observations, zero-demand periods, or extreme values.

### 3.2. Model Selectors

This section formalizes the three selectors evaluated in the study: AHS, OWA, and ERA. All selectors operate on the same observed out-of-sample evaluation space, but each applies a different decision logic. AHS is the proposed adaptive selector that uses demand-structure characterization, hierarchical metric refinement, Pareto filtering, and bias-aware tie-breaking. OWA is a benchmark-relative selector based on sMAPE and MASE, while ERA is a ranking-aggregation comparator based on MAE, RMSE, and $R^2$.

Let $M = \{m_1, m_2, \dots, m_K\}$ denote the set of candidate models evaluated on the same demand series. For each model $m \in M$, out-of-sample performance metrics and multi-step forecast predictions are available. These elements provide the basis for the selection criteria formalized below.

#### *3.2.1. Adaptive Hybrid Selector*

The Adaptive Hybrid Selector (AHS) is introduced as an adaptive model-selection procedure for structurally heterogeneous demand environments. It addresses cases in which a single monometric selection criterion may be insufficient for frequent, intermittent, highly variable, lumpy, or disaggregated multi-SKU demand series. This design is motivated by evidence that forecasting performance depends not only on the model itself, but also on demand intermittency, variability, sparsity, disaggregation, and product-level heterogeneity, particularly in multi-product settings where no universally dominant model exists (Flyckt & Lavesson, 2025; Giannopoulos et al., 2025; Hyndman & Athanasopoulos, 2021; Khan & Al Hanbali, 2025; Makridakis et al., 2018).

AHS is not interpreted as an unrestricted aggregation of heterogeneous error measures into a universal accuracy functional, since different metrics may elicit different functionals of the predictive distribution (Gneiting, 2011; Kolassa, 2020). Instead, it follows a hierarchical decision rule in which the scaled error serves as the primary selection layer, while additional indicators are used only for Pareto refinement, diagnostic interpretation, or tie-breaking. For each candidate model $m \in M$, the AHS evaluation vector is defined as:

$$e_m = (RMSSE_m, MAE_m, sMAPE_m, BIAS_m) \tag{1}$$

where $RMSSE_m$ is the primary scale-normalized selection criterion, $MAE_m$ provides auxiliary absolute-error information, and $sMAPE_m$ and $BIAS_m$ are retained as diagnostic and tie-breaking indicators. Thus, the methodological role of each metric is differentiated: $RMSSE_m$ controls scaled forecast accuracy, $MAE_m$ controls absolute deviations, $sMAPE_m$ provides relative-error diagnosis, and $BIAS_m$ captures directional systematic error. This avoids treating heterogeneous metrics as equivalent while preserving their operational relevance in inventory contexts (Gneiting, 2011; Hyndman & Koehler, 2006; Kolassa, 2020).

AHS also incorporates two operational structural descriptors of the demand series $S$: demand frequency $(FD_S)$ and the coefficient of variation of the complete series $(CVSC_S)$. These descriptors are consistent with the broader forecasting literature that characterizes demand series according to intermittency and variability, although they are used here as operational activation variables rather than as a direct replacement for the ADI–CV² taxonomy of Syntetos et al. (2005). Demand frequency is defined as the proportion of periods with positive demand:

$$FD_S = \frac{1}{T}\sum_{t=1}^{T} I(y_t > 0) \tag{2}$$

where $I(\cdot)$is an indicator function equal to 1 when demand is positive and 0 otherwise. The coefficient of variation of the complete series is defined as:

$$CVSC_S = \frac{\sigma_S}{\bar{y}_S} \tag{3}$$

where $\sigma_S$ and $\bar{y}_S$ denote the standard deviation and mean of the full demand series, respectively. In the empirical implementation, the adaptive condition was evaluated using $p^* = 0.5$ as the demand-frequency threshold and $c^* = 0.7$ as the variability threshold. These values are used as operational activation thresholds within AHS and should not be interpreted as universal demand-classification cutoffs. Given these thresholds, the adaptive condition is defined as:

$$FD_S \geq p^* \text{ and } CVSC_S < c^* \tag{4}$$

If this condition is satisfied, the series is interpreted as frequent and relatively stable, and AHS activates its adaptive hierarchical rule. Candidate models are first compared through Pareto dominance over two core minimization criteria:

$$\min\{RMSSE_m, MAE_m\} \tag{5}$$

The Pareto-efficient candidate set is defined as:

$$P(S) = \{m \in M : \nexists\, m' \in M \text{ such that } m' \succ m\} \tag{6}$$

where $m' \succ m$ means that model $m'$ dominates model $m$. This dominance relation is defined as:

$$m' \succ m \Leftrightarrow RMSSE_{m'} \leq RMSSE_m,\ MAE_{m'} \leq MAE_m, \text{ and } (RMSSE_{m'} < RMSSE_m \text{ or } MAE_{m'} < MAE_m) \tag{7}$$

Therefore, $P(S)$ contains the candidate models that are not simultaneously dominated in scaled error and absolute error. This Pareto layer prevents AHS from selecting a model solely because it performs well on a single metric while performing poorly on another relevant error dimension.

Once the Pareto-efficient set $P(S)$ is obtained, AHS refines the selection using $sMAPE_m$. The selected model is first defined as:

$$m^* = \arg\min_{m \in P(S)} sMAPE_m \tag{8}$$

If more than one model remains tied after the $sMAPE_m$-based refinement, the final decision is obtained by minimizing the absolute value of the forecast bias:

$$m^* = \arg\min_{m \in T(S)} |\, BIAS_m \,| \tag{9}$$

where $T(S)$ denotes the subset of models tied after the $sMAPE_m$-based refinement. This hierarchy preserves each metric's methodological role. $RMSSE_m$ acts as the primary scaled-error criterion, $MAE_m$ supports Pareto-based absolute-error refinement, $sMAPE_m$ provides a relative-error refinement within the efficient candidate set, and $|\, BIAS_m \,|$ acts as a directional tie-breaking indicator. The use of bias is operationally relevant because systematic overestimation or underestimation may affect inventory exposure, stockouts, replenishment stability, and service performance (Doszyń & Dudek, 2024; Zhao et al., 2023).

If the adaptive condition is not satisfied, AHS interprets the series as irregular, intermittent, or highly variable. In this case, the selector adopts a conservative fallback rule and selects the model that minimizes the scaled error over the complete candidate set:

$$m^* = \arg\min_{m \in M} RMSSE_m \tag{10}$$

This fallback rule avoids unstable multicriteria refinements when demand is sparse, intermittent, or highly variable, since percentage-based metrics, bias indicators, or Pareto comparisons may be affected by zero-demand observations, extreme values, or irregular demand occurrence. Under these conditions, $RMSSE_m$ provides a more robust scale-normalized basis for comparing candidate models across heterogeneous demand series. The complete AHS decision rule can be summarized as:

$$m^* = \begin{cases} \arg\min_{m \in P(S)} sMAPE_m, & \text{if } FD_S \geq p^* \text{ and } CVSC_S < c^* \\ \arg\min_{m \in M} RMSSE_m, & \text{otherwise} \end{cases} \quad (11)$$

with the additional tie-breaking condition:

$$m^* = \arg\min_{m \in T(S)} | BIAS_m | \quad (12)$$

Overall, AHS defines a regime-sensitive, hierarchy-based selection procedure that combines structural demand descriptors, Pareto-based refinement, relative-error diagnosis, and bias-aware tie-breaking. It operates directly on observed model-performance indicators and structural descriptors of the demand series. For frequent and relatively stable demand, it applies a hierarchical multicriteria rule; for irregular, intermittent, or highly variable demand, it applies a conservative scaled-error minimization rule. This structure is consistent with recent forecasting research on adaptive model selection, demand-structure classification, and model assignment in intermittent, lumpy, disaggregated, and multi-SKU environments (Giannopoulos et al., 2025; Khan & Al Hanbali, 2025; Shi et al., 2026; Song et al., 2025; Taghiyeh et al., 2020).

### *3.2.2. Equilibrium Ranking Aggregation Selector*

The Equilibrium Ranking Aggregation (ERA) selector is incorporated as a multivariate ranking-aggregation selector for comparing candidate forecasting models under a common out-of-sample evaluation space. ERA is motivated by inventory-oriented demand forecasting studies that assess model performance using several accuracy indicators and relate it to operational outcomes such as stockout reduction, overstock reduction, inventory turnover, and cost improvement (Kumar et al., 2025). In this sense, ERA provides a pragmatic decision rule for selecting the model that achieves the best aggregate position across complementary evaluation metrics.

ERA does not assume that MAE, RMSE, and R² represent interchangeable dimensions of a single universal accuracy concept. Instead, it treats them as complementary indicators with different methodological roles. MAE captures the absolute magnitude of forecast errors, RMSE penalizes larger deviations more strongly, and R² is retained as a descriptive indicator of explanatory capacity. For each candidate model $M_i \in M$, the ERA evaluation vector is defined as:

$$e_i = (MAE_i, RMSE_i, R_i^2) \quad (13)$$

where lower values of $MAE_i$ and $RMSE_i$ indicate better error performance, while higher values of $R_i^2$ indicate stronger explanatory capacity. Let $M = \{M_1, M_2, \dots, M_K\}$ denote the set of candidate forecasting models, and let

$$J = \{MAE, RMSE, R^2\} \quad (14)$$

Denote the set of indicators used by ERA. For each indicator, an independent ranking is constructed. Since MAE and RMSE are error metrics, they are ranked in ascending order. Since $R^2$ is a goodness-of-fit indicator, it is ranked in descending order. Thus, for each model $M_i$, the metric-specific rankings are:

$$r_{MAE}(M_i) = rank_{asc}(MAE_i) \quad (15)$$

$$r_{RMSE}(M_i) = rank_{asc}(RMSE_i) \quad (16)$$

$$r_{R^2}(M_i) = rank_{desc}(R_i^2) \quad (17)$$

The aggregate ERA ranking score is then computed as the sum of the metric-specific ranks:

$$ERA_{total}(M_i) = r_{MAE}(M_i) + r_{RMSE}(M_i) + r_{R^2}(M_i) \quad (18)$$

Lower values of $ERA_{total}(M_i)$ indicate better aggregate ranking positions. To express the result on a normalized scale, the ERA score is defined as:

$$F_{ERA}(M_i) = 1 - \frac{ERA_{total}(M_i) - \min(ERA_{total})}{\max(ERA_{total}) - \min(ERA_{total})} \quad (19)$$

Where $F_{ERA}(M_i) \in [0,1]$, and higher values indicate better aggregate performance. The selected forecasting model is therefore:

$$M^* = \arg\max_{M_i \in M} F_{ERA}(M_i) \quad (20)$$

Accordingly, ERA is defined as a coherent and reproducible ranking-aggregation selector based on MAE, RMSE, and $R^2$. Its purpose is to identify the model with the best overall ordinal position across absolute error, quadratic error, and explanatory capacity, following the inventory-forecasting logic in which predictive accuracy is evaluated through multiple indicators and connected to operational inventory performance.

### *3.2.3. Overall Weighted Average*

The Overall Weighted Average (OWA) selector is incorporated as a benchmark-relative model-selection criterion based on two scale-independent forecast accuracy measures: $sMAPE$ and $MASE$. Its use is motivated by automatic demand forecast model-selection studies in which $sMAPE$, $MASE$, and $OWA$ are used to evaluate and select forecasting models at the individual time-series level. In Garred et al. (2026), OWA is computed by comparing each candidate model against a benchmark model, using the relative performance of $sMAPE$ and $MASE$; the model with the best OWA value is then selected for forecasting future demand.

For a demand series $S$, let $M = \{m_1, m_2, \ldots, m_K\}$ denote the set of candidate forecasting models, and let $B = \{b_1, b_2, \ldots, b_L\}$ denote an ordered set of benchmark models. For each candidate model $m \in M$, the OWA evaluation vector is defined as:

$$e_m = (sMAPE_m, MASE_m) \quad (21)$$

where $sMAPE_m$ measures relative percentage error and $MASE_m$ provides a scale-independent error measure based on naïve-error scaling. Since both metrics are minimization criteria, lower values indicate better forecasting performance.

OWA requires a valid benchmark model $b \in B$. The benchmark must be available for the series $S$, and its error values must be strictly positive:

$$sMAPE_b > 0 \tag{22}$$

$$MASE_b > 0 \tag{23}$$

If no valid benchmark is available, OWA cannot be computed for that demand series because the relative normalization would be undefined. In the empirical implementation, OWA was computed using the Seasonal Naïve model as the benchmark, following Garred et al. (2026). Therefore, each candidate model was compared with Seasonal Naïve using relative sMAPE and MASE ratios. Accordingly, OWA values below 1 indicate that the candidate model outperforms the Seasonal Naïve benchmark, whereas values above 1 indicate that Seasonal Naïve provides better accuracy. Once a valid benchmark $b$ is identified, each candidate model is compared against it through relative error ratios. The relative $sMAPE$ is defined as:

$$r_{sMAPE,m} = \frac{sMAPE_m}{sMAPE_b} \tag{24}$$

and the relative $MASE$ is defined as:

$$r_{MASE,m} = \frac{MASE_m}{MASE_b} \tag{25}$$

These ratios indicate whether the candidate model improves upon the benchmark. A value below 1 indicates that the candidate model performs better than the benchmark on that metric, while a value above 1 indicates that the benchmark performs better. The OWA score of model $m$ is then computed as the arithmetic mean of both relative errors:

$$OWA_m = \frac{1}{2}\left(r_{MASE,m} + r_{sMAPE,m}\right) \tag{26}$$

Equivalently:

$$OWA_m = \frac{1}{2}\left(\frac{MASE_m}{MASE_b} + \frac{sMAPE_m}{sMAPE_b}\right) \tag{27}$$

Thus, OWA combines the benchmark-relative performance of $MASE$ and $sMAPE$ into a single comparative score. Since both components are error ratios, lower OWA values indicate better aggregate performance relative to the benchmark. The selected forecasting model is therefore:

$$m^* = \arg\min_{m \in M} OWA_m \tag{28}$$

Accordingly, the OWA selector chooses the candidate model with the lowest average relative error relative to the selected benchmark. Its main advantage is that it does not rely on a single accuracy metric; instead, it evaluates each model through the joint relative behavior of $sMAPE$ and $MASE$.

This makes OWA useful as a benchmark-relative selector in heterogeneous demand-forecasting contexts, where model performance may differ depending on whether percentage error or scaled error is emphasized.

### 3.3. Global Relative Accuracy as an Ex Post Volumetric Indicator

From a business perspective, demand forecasts are not only statistical outputs but also inputs for purchasing, replenishment, and inventory-planning decisions. In these settings, the total forecasted demand over an operational evaluation horizon may be interpreted as the volume that the organization anticipates, allocates, or commits in advance to serve future demand. Therefore, beyond pointwise error behavior, it is useful to assess whether the model selected by each selector produces an aggregate forecast volume that is coherent with the realized demand volume across the evaluated future cycles (Kumar et al., 2025; Nguyen, 2026; Sayed et al., 2009; Sepúlveda-Rojas et al., 2015; Ulrich et al., 2022).

For this purpose, the Global Relative Accuracy (GRA) indicator is used as an ex post measure of volumetric coherence between total forecasted demand and total realized demand. It is defined as:

$$GRA = 1 - \frac{| \sum_{t=1}^{T} \hat{y}_t - \sum_{t=1}^{T} y_t |}{\sum_{t=1}^{T} y_t} \tag{29}$$

where $y_t$ denotes observed demand, $\hat{y}_t$ denotes predicted demand, and $T$ is the number of future cycles included in the evaluation horizon. Values closer to 1 indicate higher aggregate agreement between forecasted and realized demand volumes. Because GRA is defined as one minus a relative aggregate deviation, values may become negative when the absolute volumetric error exceeds the realized demand volume. Therefore, GRA is interpreted as an ex post volumetric alignment indicator rather than a bounded-accuracy score. To avoid numerical instability, cases with zero realized aggregate demand over the evaluation horizon are excluded from GRA computation.

GRA is not used as a primary criterion for model selection. Instead, it is applied after model selection to evaluate whether the selected models remain volumetrically consistent across the future cycles considered in the empirical assessment. In this sense, GRA provides an operational validation layer: it measures whether the demand volume implied by the selected forecasts is close to the realized aggregate demand volume. Because GRA is a relative aggregate indicator, it may inherit limitations associated with percentage-based measures if interpreted as a statistical accuracy criterion; therefore, it is used only to evaluate aggregate volumetric alignment, not to elicit the target functional of the point forecast (Gneiting, 2011; Kolassa, 2020; Kolassa & Schütz, 2007).

Operationally, higher GRA values indicate better alignment between selected forecasts and realized demand volume across the evaluated cycles. Conversely, persistent aggregate overestimation or underestimation may indicate that the selected model is not operationally consistent with future demand volume, even if it performs adequately under pointwise error metrics. This distinction is relevant in inventory contexts, since aggregate forecast imbalance may lead to overstock, immobilized capital, stockouts, lost sales, or service deterioration (Demizu et al., 2023; Goltsos et al., 2022; Nguyen, 2026; Ulrich et al., 2022).

However, GRA does not explicitly model the economic consequences of forecast errors. It does not distinguish between holding costs, shortage costs, service-level penalties, replenishment constraints, or production-capacity restrictions. Therefore, the present study uses GRA as an intermediate operational indicator of volumetric coherence rather than as a full inventory-cost or production-planning objective. A complete economic validation would require integrating the selected forecasts into inventory-control or production-planning simulations with explicit cost parameters, service-level targets, and replenishment policies. This extension is beyond the scope of the current study and is identified as a relevant direction for future research.

Thus, GRA is used to compare the ex post volumetric consequences of the evaluated selectors and to assess whether their selected models produce demand forecasts that remain consistent with realized aggregate demand across future cycles. It does not define or optimize the primary forecasting function, nor does it represent the final economic performance of an inventory system.

### 3.4. Backtesting Strategy

For each demand series, the available historical observations are represented as a discrete time series $\{y_t\}_{t=1}^{T}$, where $y_t$ denotes observed demand at period $t$ and $T$ is the total number of observations. The last $H$ cycles are reserved as an out-of-sample evaluation segment, while the preceding observations form the estimation segment. The notion of cycle is defined according to the intrinsic frequency of each series, whether daily, weekly, monthly, or annual; therefore, $H$ represents a multi-step forecasting horizon aligned with the corresponding operational planning unit.

The adopted design corresponds to a fixed-origin out-of-sample evaluation framework. Each candidate model is trained only on observations prior to the evaluation horizon and then generates multi-step forecasts for the reserved segment, which is excluded from model estimation and hyperparameter fitting. After the forecasting process, the predictions produced by the selected model are compared with the realized values in the reserved segment. This comparison provides the empirical basis for computing out-of-sample error metrics and for evaluating whether the selected model remains consistent with the observed demand over the future cycles considered in the assessment.

This strategy preserves temporal ordering, avoids information leakage, and is consistent with recommended practices for out-of-sample and multi-step forecasting assessment (Hyndman & Koehler, 2006; Kourentzes et al., 2014; Semenoglou et al., 2021; Tashman, 2000). It is also appropriate for demand forecasting contexts, where predictions support operational planning, demand and supply decisions, inventory management, and resource allocation (Armstrong, 2001; Makridakis et al., 2022).

### 3.5. Experimental Design

To evaluate automatic model selection in heterogeneous demand forecasting environments, the experimental framework compares three selectors operating on a common out-of-sample evaluation space: AHS, ERA, and OWA. AHS represents the proposed adaptive selector, ERA provides a multivariate ranking-aggregation rule, and OWA acts as a benchmark-relative selector based on sMAPE and MASE. The workflow uses the same candidate models, temporal validation scheme, observed performance metrics, and an ex post GRA assessment of volumetric coherence. Figure 1 summarizes the experimental design in 10 stages grouped into four blocks: data preparation, model estimation, selector-based assignment, and ex post forecast assessment.

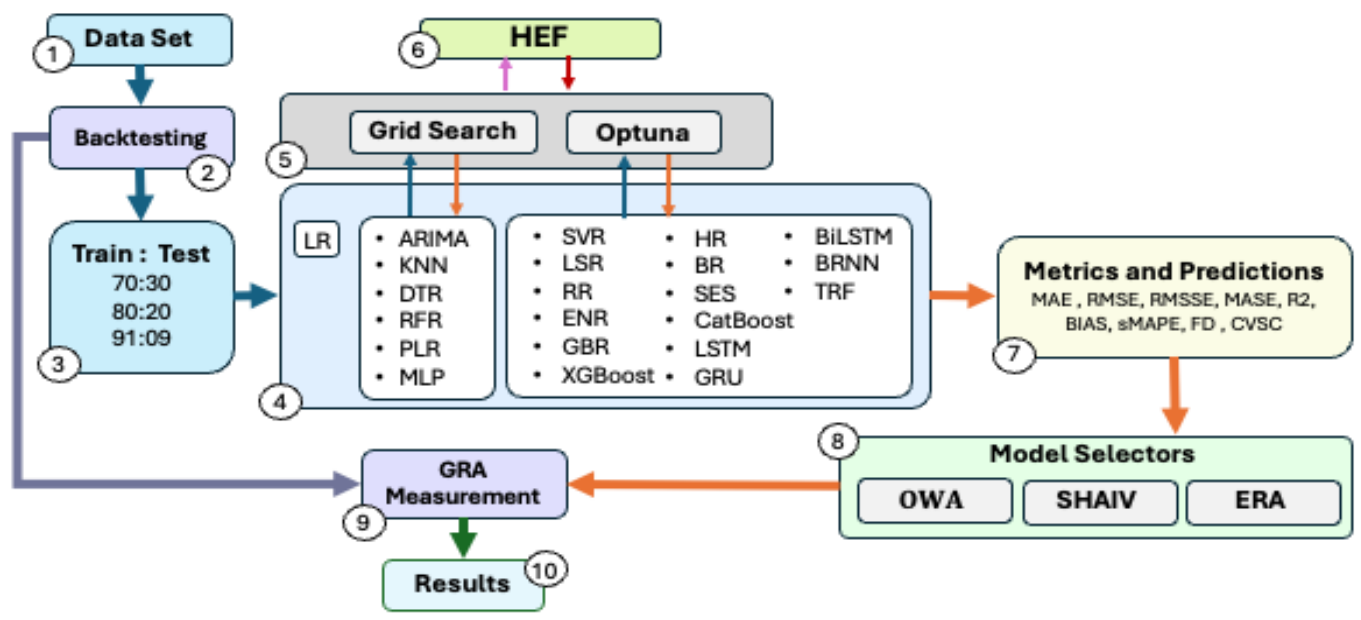

**Figure 1.** Experimental Framework.

### *3.5.1. Stage 1: Dataset.*

The empirical analysis uses four widely recognized forecasting datasets: Walmart (Yasser, 2021), M3 (Makridakis & Hibon, 2000), M4 (Makridakis et al., 2018), and M5 (Makridakis et al., 2022). These datasets were selected because they represent heterogeneous demand scenarios with varying non-stationarity, intermittency, variability, and structural change, thereby providing a realistic setting for evaluating model selection under increasing structural complexity. Table 1 reports the main characteristics of the original datasets, while Table 2 presents the stratified random samples used in the experiments. The sample size was determined under a finite-population design with a 99% confidence level, a 5% margin of error, and an expected proportion of $p = 0.5$.

In addition to the general dataset descriptors, the demand-pattern classes reported in Tables 1 and 2 were obtained using the classification scheme proposed by Syntetos et al. (2005). This taxonomy classifies each individual time series according to two structural indicators: the average inter-demand interval, denoted as ADI, and the squared coefficient of variation of positive demand sizes, denoted as $CV^2$. Following this scheme, each series was assigned to one of four demand-pattern classes: smooth demand, when ADI is lower than 1.32 and $CV^2$ is lower than 0.49; intermittent demand, when ADI is equal to or greater than 1.32 and $CV^2$ is lower than 0.49; erratic demand, when ADI is lower than 1.32 and $CV^2$ is equal to or greater than 0.49; and lumpy demand, when ADI is equal to or greater than 1.32 and $CV^2$ is equal to or greater than 0.49. The resulting classifications were computed at the individual-series level and subsequently aggregated to characterize each original dataset and experimental sample. This descriptive taxonomy should be distinguished from the operational descriptors used by AHS, namely demand frequency (FD) and the coefficient of variation of the series (CVSC), which activate the adaptive selection logic within the proposed selector.

**Table 1.** Datasets used in the experiments.

| Dataset | Frequency | # Time Series | # Observations | Annual Frequency | Classes |
|---|---|---|---|---|---|
| Walmart | Weekly | 45 | 4,680 | 52 | Smooth (100%) |
| M5 | Daily | 1,454 | 1,054,150 | 365 | Smooth (9.2%), Intermittent (76.1%), Erratic (2.2%), Lumpy (12.5%) |
| M4 | Weekly | 294 | 28,224 | 52 | Smooth (99%), Erratic (1%) |
| M3 | Monthly | 1,428 | 34,272 | 12 | Smooth (99%), Erratic (1%) |

| Total | - | 3,221 | 1,121,326 | - | - |
|---|---|---|---|---|---|

**Table 2.** Stratified random sample used in the experiments.

| Dataset | Frequency | # Time Series | # Observations | Annual Frequency | Classes |
|---|---|---|---|---|---|
| Walmart | Weekly | 45 | 4,680 | 52 | Smooth (100%) |
| M5 | Daily | 650 | 471,250 | 365 | Smooth (9.2%), Intermittent (76.3%), Erratic (2.3%), Lumpy (12.2%) |
| M4 | Weekly | 204 | 19,584 | 52 | Smooth (99.1%), Erratic (0.9%) |
| M3 | Monthly | 454 | 10,896 | 12 | Smooth (99%), Erratic (1%) |
| **Total** | **-** | **1,351** | **506,410** | **-** | **-** |

### *3.5.2. Stage 2: Backtesting.*

A fixed-origin temporal backtesting scheme was adopted to preserve the chronological structure of each series and prevent leakage of future information (Hyndman & Koehler, 2006; Tashman, 2000). For all datasets, the future evaluation horizon was set to $H = 12$ cycles, with the notion of cycle adapted to the intrinsic frequency of each dataset. Accordingly, the evaluation horizon is 12 weeks for the weekly series, 12 months for the monthly series, and 12 daily periods for the daily series. This design ensures comparability across datasets while remaining consistent with the operational planning unit of each series.

### *3.5.3. Stage 3: Training and test separation.*

Each sampled dataset was evaluated under three training–testing configurations: 91:9, 80:20, and 70:30. These partition schemes allow examination of selector behavior under different levels of historical data availability while keeping the evaluation protocol consistent across all experiments.

### *3.5.4. Stage 4: Demand forecasting models.*

The candidate forecasting set comprises 22 models, including traditional statistical methods, machine learning regressors, ensemble-based methods, and deep learning architectures. Specifically, the evaluated models were ARIMA, KNN, DTR, RFR, PLR, MLP, LR, SVR, LSR, RR, ENR, GBR, XGBoost, HR, BR, SES, CatBoost, LSTM, GRU, BiLSTM, BRNN, and TRF. As detailed in Table 3, these models were organized into two groups: Exhaustive Search (ES), for methods optimized over finite discrete spaces, and Search in Continuous Space (SCS), for models requiring continuous hyperparameter optimization in higher-dimensional spaces.

**Table 3.** Candidate forecasting models and optimization groups.

| Model | Cluster | Explanation |
|---|---|---|
| ARIMA | ES | Autoregressive Integrated Moving Average is a classical time series approach that captures linear relationships through autoregressive components, moving averages, and differencing to address trends. It is particularly effective for stationary data or for series that can be rendered stationary through appropriate transformations (Box et al., 2015). |
| KNN | ES | K-Nearest Neighbors is a nonparametric technique that estimates values based on the average of the k nearest neighbors in the feature space. It is characterized by its simplicity, interpretability, and its capacity to handle nonlinear data structures (Altman, 1992). |
| DTR | ES | Decision Tree Regression generates partitions through hierarchical decision rules, resulting in interpretable models, although they are prone to overfitting (Breiman et al., 2017). |
| RFR | ES | Random Forest Regressor constructs an ensemble of multiple decision trees trained on random subsets of the data, thereby improving predictive accuracy and reducing overfitting (Breiman, 2001). |
| PLR | ES | Polynomial Regression extends linear regression by incorporating polynomial terms, thereby enabling the modeling of nonlinear relationships in a straightforward manner (Montgomery et al., 2021). |

| MLP | ES | Multilayer Perceptron is an artificial neural network with one or more hidden layers, capable of capturing complex nonlinear relationships and widely used in regression and classification tasks (Haykin, 1994). |
|---|---|---|
| LR | - | Linear Regression is a fundamental regression model that estimates the relationship between independent and dependent variables by minimizing the squared error (Montgomery et al., 2021). |
| SVR | SCS | Support Vector Regression is an extension of support vector machines for regression tasks, designed to identify a function that lies within a specified tolerance margin from the observed values, employing kernel functions to capture complex relationships (Smola & Schölkopf, 2004). |
| LSR | SCS | Lasso Regression incorporates an L1 regularization penalty, thereby facilitating automatic variable selection by shrinking less relevant coefficients to zero (Tibshirani, 1996). |
| RR | SCS | Ridge Regression applies an L2 regularization penalty, which mitigates multicollinearity without eliminating variables (Hoerl & Kennard, 1970). |
| ENR | SCS | Elastic Net Regression combines L1 and L2 penalties, enabling the selection of relevant variables while controlling for collinearity (Zou & Hastie, 2005). |
| GBR | SCS | Gradient Boosting Regressor constructs decision trees sequentially, where each new tree corrects the errors of the preceding ones, achieving high predictive performance at the cost of increased calibration complexity (Friedman, 2001). |
| XGBoost | SCS | An optimized implementation of gradient boosting improves computational speed, regularization, and the handling of missing data, making it particularly effective for structured prediction problems (Chen & Guestrin, 2016). |
| HR | SCS | Huber Regressor is a robust regression technique that employs the Huber loss function, combining sensitivity to small squared errors with resistance to outliers (P. J. Huber, 1964). |
| BR | SCS | Bayesian Ridge Regression introduces prior distributions over the regression coefficients, thereby incorporating uncertainty into the parameter estimates (MacKay, 1992). |
| SES | SCS | Simple Exponential Smoothing is applied to time series without trend or seasonality, assigning greater weight to recent observations through an exponential decay rule (Brown, 1959). |
| CatBoost | SCS | A boosting algorithm that efficiently handles categorical variables without explicit encoding and reduces overfitting through ordered boosting (Prokhorenkova et al., 2018). |
| LSTM | SCS | Long Short-Term Memory is a recurrent architecture designed to capture long-term dependencies in sequential data, particularly useful in time series applications (Hochreiter & Schmidhuber, 1997). In the present study, a regression-oriented variant is implemented, consisting of two LSTM layers (with 128 and 64 units, respectively, and tanh activation), a dropout mechanism of 30%, an intermediate dense layer with 32 neurons (ReLU activation), and a linear output layer. The model is compiled using the Adam optimizer and employs mean squared error (MSE) as the loss function. |
| GRU | SCS | Gated Recurrent Unit is a recurrent architecture designed to model temporal dependencies through a simplified gating mechanism, reducing computational complexity compared to LSTM while maintaining an adequate capacity to represent sequential dynamics, which makes it particularly suitable for time series applications (Cho et al., 2014). In the present study, a regression-oriented variant is implemented, consisting of a single GRU layer with 64 units and tanh activation, followed by an intermediate dense layer with 64 neurons and ReLU activation. The output layer corresponds to a dense layer whose dimensionality is adjusted according to the prediction horizon under consideration, thereby enabling direct multi-step forecasting. The model is compiled using the Adam optimizer and employs mean squared error (MSE) as the loss function, ensuring consistency in predictive performance evaluation. |
| BiLSTM | SCS | Bidirectional Long Short-Term Memory Neural Network is a bidirectional recurrent architecture that extends the LSTM model by simultaneously processing the temporal sequence in both forward and backward directions, thereby enabling the capture of past and future dependencies within the observation window and improving the representation of complex temporal patterns (Graves & Schmidhuber, 2005; Schuster & Paliwal, 1997). In the present study, a multi-step regression-oriented variant is implemented, consisting of two stacked BiLSTM layers with 64 and 32 units, respectively. The first layer operates in full sequential mode, while the second condenses the temporal information into a final state representation. To mitigate the risk of overfitting, dropout regularization with a rate of 20% is applied after each recurrent layer. The resulting representation is subsequently processed through two intermediate dense layers with 64 and 32 neurons, respectively, both using ReLU activation, before applying a dense output layer with linear activation whose dimensionality is adjusted according to the prediction horizon under consideration, thereby enabling direct multi-step forecasting. The model is trained using the Adam optimizer and employs mean squared error (MSE) as the loss function, ensuring consistency in predictive performance evaluation. |
| BRNN | SCS | Bidirectional Simple Recurrent Neural Network is a bidirectional recurrent architecture based on simple recurrent units, which processes the input temporal sequence simultaneously in forward and backward directions, thereby capturing bidirectional temporal dependencies within the observation window, albeit with a simpler structure than LSTM or GRU variants (Schuster & Paliwal, 1997). In the present study, a multi-step regression-oriented variant is implemented, consisting of a single bidirectional SimpleRNN layer with 64 units and tanh activation, which condenses the temporal information into a final representative state. The resulting representation is subsequently processed through an intermediate dense layer with 64 neurons and ReLU activation, before applying a dense output layer whose dimensionality is adjusted according to the prediction horizon under consideration, thereby enabling direct multi-step forecasting. The model is trained using the Adam optimizer and employs mean squared error (MSE) as the loss function, ensuring consistent evaluation of predictive performance. |
| TRF | SCS | Transformer Model for Time Series is an architecture based on autoregressive self-attention mechanisms that dispenses with explicit recurrence and enables the modeling of long-range temporal dependencies through multi-head attention, making it particularly suitable for capturing complex and non-local relationships in temporal sequences (Vaswani et al., 2017). In the present study, a multi-step regression-oriented variant is implemented, beginning with a dense projection of the input into a 64-dimensional representation space, followed by a Transformer block composed of a multi-head attention mechanism with two heads and an internal feed-forward network of 128 units. The block incorporates residual connections, layer normalization, and dropout regularization, allowing stable information propagation and robust learning of temporal dependencies. The resulting representation is subsequently transformed through flattening and two intermediate dense layers with 64 and 32 neurons, respectively, both using ReLU activation, before applying a dense output layer with linear activation whose dimensionality is adjusted according to the prediction horizon under consideration, thereby enabling direct multi-step forecasting. The model is trained using the Adam optimizer and employs mean squared error (MSE) as the loss function, ensuring consistency in predictive performance evaluation. |

### *3.5.5. Stage 5: Optimizers.*

Two optimization strategies were applied based on the model structure. Grid Search was used for ES models to ensure complete exploration of the predefined discrete parameter space. Optuna was used for SCS models, enabling efficient hyperparameter optimization in continuous spaces via Bayesian search and pruning.

### *3.5.6. Stage 6: HEF.*

All model parameters, hyperparameters, and fixed configuration settings were calibrated or defined using the Hierarchical Evaluation Function (HEF) (González & Parada, 2026) as a common optimization criterion. HEF was used to ensure methodological consistency across models by integrating explanatory power, average error magnitude, and sensitivity to large errors within a unified evaluation structure. This reduces the risk that subsequent selector comparisons reflect differences in optimization criteria rather than genuine differences in predictive performance. Table 4 presents the model-specific search spaces and fixed settings used for each model during training and optimization.

**Table 4.** Parameter ranges explored for the models considered in this study.

| Model | Parameters |
|---|---|
| ARIMA | p=0-5, d=0-3, q=0-5 |
| KNN | n_neighbors between 1 and 35 |
| DTR | max_depth between 1 and 32; min_samples_split = 2; min_samples_leaf = 1; random_state = 21 |
| RFR | n_estimators between 1 and 100; random_state = 21 |
| PLR | degree between 1 and 3; include_bias = False; interaction_only = True |
| MLP | hidden_layer_sizes = [layer 1: 16–32; layer 2: 0–32; layer 3: 0–32]; activation = 'relu'; solver = 'adam'; max_iter = 1000; early_stopping = True; n_iter_no_change = 10; random_state = 21 |
| LR | Defaults |
| SVR | kernel = 'rbf'; C between 0.01 and 20; degree between 2 and 5; epsilon between 0.01 and 1; gamma = 'scale'. |
| LSR | alpha between 0.01 and 5; random_state = seed_value; max_iter = 8000. |
| RR | alpha between 0.1 and 10; max_iter = 8000 |
| ENR | alpha between 0.01 and 0.02; l1_ratio between 1e-3 and 0.1 |
| GBR | n_estimators between 10 and 2000; learning_rate between 0.1 and 0.3; random_state = seed_value |
| XGBoost | n_estimators between 10 and 2000; learning_rate between 0.1 and 0.3; max_depth between 3 and 5; enable_categorical = True; random_state = seed_value |
| HR | epsilon between 1 and 3; max_iter = 1000; alpha = 0.0001; tol = 1e-4 |
| BR | max_iter between 50 and 1000; alpha_1 between 1e-6 and 1e-2; alpha_2 between 1e-6 and 1e-2 |
| SES | smoothing_level between 0.01 and 0.9; optimized = False |
| CatBoost | iterations between 100 and 1000; depth between 3 and 10; learning_rate between 0.01 and 0.3; loss_function = 'Huber:delta=1.0'; train_dir = tmp_dir; verbose = False; random_seed = 21 |
| LSTM | epochs between 1 and 80 |
| GRU | epochs between 1 and 80 |
| BiLSTM | epochs between 1 and 80 |
| BRNN | epochs between 1 and 80 |
| TRF | epochs between 1 and 80 |

The optimization was addressed through two distinct approaches. In Optuna, each training and optimization cycle was conducted within a fixed budget of 21 trials, enabling efficient exploration of the search space via Bayesian techniques. In contrast, Grid Search was implemented via exhaustive parameter sweeps, systematically evaluating all possible combinations within the predefined discrete search spaces.

#### *3.5.7. Stage 7: Metrics and predictions.*

Once trained under the corresponding partition scheme, the models generated multi-step forecasts for horizons h = 1, 2, …, 12. Then, MAE, RMSE, RMSSE, MASE, $R^2$, BIAS, sMAPE, FD, and CVSC were computed on the training and testing segments, while future predictions were generated without introducing future information into the estimation process. These outputs constitute the basis for the selector comparison stage.

#### *3.5.8. Stage 8: Model selectors.*

Based on the observed out-of-sample metrics produced in the previous stage, three selector mechanisms were applied to each demand series. Within the methodological architecture of the study, AHS operates as the proposed adaptive selector, integrating demand structure and hierarchical metric refinement; ERA acts as a multivariate ranking-aggregation selector; and OWA provides a benchmark-relative selection rule based on sMAPE and MASE. Algorithms 1–3 summarize their operational implementation within the experimental process, following the formal definitions presented above.

**Algorithm 1.** Adaptive Hybrid Selector (AHS)

```
ALGORITHM AHS (S, M)

INPUT:
  S      → Time series or demand series to be forecasted.
  M      → Set of candidate forecasting models.

  For each model m ∈ M:
    RMSSE_m  → Root mean squared scaled error.
    MAE_m    → Mean absolute error.
    sMAPE_m  → Symmetric mean absolute percentage error.
    BIAS_m   → Forecast bias.

  For the series S:
    FD_S     → Demand frequency.
    CVSC_S   → Coefficient of variation of the series.
```

```
PARAMETERS:
   p*     → Demand-frequency threshold.
   c*     → Variability threshold.

OUTPUT:
   m*     → Selected forecasting model.

BEGIN
   # STRUCTURAL CHARACTERIZATION OF THE SERIES
   Compute FD_S
   Compute CVSC_S

   # REGIME-ADAPTIVE MODEL SELECTION
   IF FD_S ≥ p* AND CVSC_S < c* THEN

      # Frequent and relatively stable demand regime
      Construct the Pareto-efficient candidate set:
         P ← Pareto-efficient models in M
             with respect to minimizing:
                RMSSE_m
                MAE_m

      Refine the Pareto candidate set:
         m* ← model in P with the lowest sMAPE_m

      If more than one model remains tied:
         m* ← model among tied candidates with the lowest |BIAS_m|

   ELSE
      # Irregular, intermittent, or highly variable demand regime
      m* ← model in M with the lowest RMSSE_m

   END IF

   RETURN m*

END
```

**Algorithm 2.** Overall Weighted Average (OWA)

---

```
ALGORITHM OWA-BASED MODEL SELECTION (S, M, B)

INPUT:
   S      → Demand series or forecasting task.
   M      → Set of candidate forecasting models.
   B      → Seasonal Naïve benchmark model.

   For each model m ∈ M:
      sMAPE_m  → Symmetric mean absolute percentage error.
      MASE_m   → Mean absolute scaled error.

   For each benchmark b ∈ B:
      sMAPE_b  → sMAPE of benchmark b.
      MASE_b   → MASE of benchmark b.

OUTPUT:
   m*      → Selected forecasting model.
   OWA_m   → Overall weighted average score for each candidate model.

BEGIN
   # BENCHMARK IDENTIFICATION
   Select the first benchmark b ∈ B such that:
      b is available for series S,
      sMAPE_b > 0,
      MASE_b > 0.

   IF no valid benchmark exists THEN:
      OWA cannot be computed for S.
      RETURN no selected model by OWA.
   END IF

   # RELATIVE ERROR NORMALIZATION
   FOR each candidate model m ∈ M DO

      Compute relative sMAPE:
```

```
      r_sMAPE_m ← sMAPE_m / sMAPE_b

    Compute relative MASE:
      r_MASE_m ← MASE_m / MASE_b

    # OVERALL WEIGHTED AVERAGE
    Compute:
      OWA_m ← 0.5 × (r_MASE_m + r_sMAPE_m)

  END FOR

  # MODEL SELECTION
  Select:
    m* ← arg min_{m ∈ M} OWA_m

  RETURN m*, OWA_m

END
```

**Algorithm 3.** Equilibrium Ranking Aggregation (ERA)

```
ALGORITHM ERA (M)

INPUT:
  M → Set of candidate forecasting models.

  For each model m ∈ M:
    MAE_m   → Mean absolute error.
    RMSE_m  → Root mean squared error.
    R²_m    → Coefficient of determination.

OUTPUT:
  m*      → Selected forecasting model.
  F_ERA_m → ERA score for each candidate model.

BEGIN
  # METRIC-SPECIFIC RANKING
  For each model m ∈ M:
    r_MAE_m  ← rank of MAE_m in ascending order
    r_RMSE_m ← rank of RMSE_m in ascending order
    r_R²_m   ← rank of R²_m in descending order

  # RANK AGGREGATION
  For each model m ∈ M:
    ERA_total_m ← r_MAE_m + r_RMSE_m + r_R²_m

  # SCORE NORMALIZATION
  For each model m ∈ M:
    F_ERA_m ← 1 −
         (ERA_total_m − min(ERA_total)) /
         (max(ERA_total) − min(ERA_total))

  # MODEL SELECTION
  m* ← arg max_{m ∈ M} F_ERA_m

  RETURN m*, F_ERA

END
```

From a computational perspective, ERA and OWA exhibit time complexity $O(K \log K)$ and $O(K)$, respectively. ERA requires ranking the $K$ candidate models across MAE, RMSE, and $R^2$, whereas OWA only computes benchmark-relative ratios for $sMAPE$ and $MASE$before selecting the model with the lowest aggregate score. AHS exhibits worst-case complexity $O(K^2)$ when the Pareto-efficient candidate set is constructed through pairwise dominance comparisons among candidate models. In the present experimental design, $K = 22$, so the decision-layer cost remains limited in absolute terms compared with the preceding stages of model estimation, hyperparameter optimization, and forecast generation. Nevertheless, this cost may become more relevant in large-scale production environments involving many candidate models, thousands of SKUs, or repeated re-selection processes.

For this reason, AHS should be interpreted as an adaptive decision-support layer whose computational cost is incurred after model training and prediction, rather than as an additional forecasting model. OWA and ERA provide complementary comparator selectors: OWA evaluates benchmark-relative performance using $sMAPE$ and $MASE$, while ERA aggregates ordinal rankings across MAE, RMSE, and $R^2$. Future research should include explicit runtime comparisons among AHS, OWA, ERA, and simpler monometric selectors, reporting selector-level execution time separately from model-training time. Such analysis would allow the marginal improvement in forecast consistency and volumetric alignment to be evaluated against the additional computational cost of each selection mechanism.

#### *3.5.9. Stage 9: GRA measurement.*

After the model assignment stage, the predictions from the selected models were compared ex post with the observed future demand values reserved during the backtesting stage. Based on this comparison, the Global Relative Accuracy (GRA) indicator was computed as an ex post aggregate measure of volumetric coherence between forecasted and realized demand. GRA was calculated for all horizons from 1 to 12 and for each training–testing partition scheme, thereby allowing comparison of selector performance across short-, medium-, and long-term scenarios.

#### *3.5.10. Stage 10: Results.*

The final stage compares selector performance across datasets, evaluation horizons, and partition schemes. The analysis combines GRA-based descriptive summaries, distributional plots, Shapiro–Wilk normality tests, Friedman tests, paired Wilcoxon post-hoc comparisons with Holm correction, and aggregated selector rankings across horizons. These outputs assess the robustness, consistency, and operational relevance of AHS, OWA, and ERA under heterogeneous demand conditions.

### 3.6. Experimental Protocol

An experimental execution protocol was defined to evaluate the proposed methodology under all combinations of datasets and training–testing partitions considered in the study. Each execution corresponds to one dataset (Walmart, M3, M4, or M5) and one partition scheme (91:9, 80:20, or 70:30), yielding a total of 12 experimental configurations. For every configuration, the same methodological pipeline was applied: data preparation, fixed-origin backtesting, model training and optimization, HEF-based calibration, multi-step forecast generation, computation of observed out-of-sample metrics, model selection using AHS, OWA, and ERA, and final ex post evaluation through GRA. This protocol ensures comparability and reproducibility across heterogeneous demand scenarios. The full execution matrix is reported in Table 5.

**Table 5.** Experimental execution matrix.

| ID | Dataset | Split | Backtesting | Model training | HEF calibration | AHS | ERA | OWA | GRA evaluation |
|---|---|---|---|---|---|---|---|---|---|
| 1 | Walmart | 91:9 | Yes | Yes | Yes | Yes | Yes | Yes | Yes |
| 2 | Walmart | 80:20 | Yes | Yes | Yes | Yes | Yes | Yes | Yes |
| 3 | Walmart | 70:30 | Yes | Yes | Yes | Yes | Yes | Yes | Yes |
| 4 | M3 | 91:9 | Yes | Yes | Yes | Yes | Yes | Yes | Yes |
| 5 | M3 | 80:20 | Yes | Yes | Yes | Yes | Yes | Yes | Yes |
| 6 | M3 | 70:30 | Yes | Yes | Yes | Yes | Yes | Yes | Yes |
| 7 | M4 | 91:9 | Yes | Yes | Yes | Yes | Yes | Yes | Yes |
| 8 | M4 | 80:20 | Yes | Yes | Yes | Yes | Yes | Yes | Yes |
| 9 | M4 | 70:30 | Yes | Yes | Yes | Yes | Yes | Yes | Yes |

| | | | | | | | | |
|---|---|---|---|---|---|---|---|---|
| 10 | M5 | 91:9 | Yes | Yes | Yes | Yes | Yes | Yes | Yes |
| 11 | M5 | 80:20 | Yes | Yes | Yes | Yes | Yes | Yes | Yes |
| 12 | M5 | 70:30 | Yes | Yes | Yes | Yes | Yes | Yes | Yes |

Each execution corresponds to one dataset and one training–testing split. The same experimental pipeline was applied in all cases: fixed-origin backtesting, model training, HEF-based calibration, multi-step forecast generation, observed out-of-sample metric computation, application of the AHS, OWA, and ERA selectors, and final ex post evaluation using GRA.

### 3.7. Hardware and Software

The forecasting models were implemented in Python using a standard scientific computing and machine learning environment. Statistical time-series models were developed with Statsmodels (Seabold & Perktold, 2010), while regression and machine learning models were implemented with Scikit-learn (Pedregosa et al., 2011). Gradient-boosting implementations were developed using XGBoost and CatBoost (Chen & Guestrin, 2016; Prokhorenkova et al., 2018), whereas neural-network-based architectures were implemented with TensorFlow, Keras and PyTorch (Abadi et al., 2016; Chollet, 2015; Paszke et al., 2019).

Data manipulation and analysis were carried out using Pandas, NumPy, and SciPy (Harris et al., 2020; The pandas development team, 2026; Virtanen et al., 2020), while Matplotlib and Seaborn were used for visualization (Hunter, 2007; Waskom, 2021). Hyperparameter optimization was performed using Grid Search and Optuna, depending on the structural characteristics of each model (Akiba et al., 2019).

The computational experiments were conducted using Jupyter Notebooks within a Conda environment, employing Python version 3.13.9 on a Linux Ubuntu operating system. The experiments were run on a 64-bit x86 architecture using an AMD Ryzen™ AI 9 HX 370 processor, with 12 physical cores, 24 logical threads, and 32 GB of LPDDR5X RAM. The system also included an NVIDIA® GeForce RTX™ 5070 Laptop GPU with 8 GB of GDDR7 memory, integrated AMD Radeon™ Graphics, and an AMD XDNA™ neural processing unit with up to 50 TOPS. Storage was provided by a 2 TB M.2 NVMe™ PCIe® 4.0 SSD. All computational experiments were executed under the same software environment to ensure consistency and reproducibility of the results.

## 4. Results

This section reports the empirical results obtained from comparing the AHS, OWA, and ERA selectors across the Walmart, M5, M4, and M3 datasets, considering forecasting horizons $h \in \{1,2,\dots,12\}$ and three training–testing partitions: 91:9, 80:20, and 70:30. The comparison is based on the ex post Global Relative Accuracy (GRA), interpreted as an operational indicator of aggregate volumetric coherence between selected forecasts and realized demand. Therefore, the results should be interpreted as downstream consequences of the model assignments produced by each selector, not as evidence that GRA defines the primary statistical target of the forecasting models. Detailed numerical summaries and statistical results supporting the findings discussed in this section are reported in Appendix A.

Overall, the evidence shows a consistent separation between ERA and the two leading selectors, AHS and OWA. ERA is generally the weakest alternative across datasets, especially in M5, M4, and M3, where its GRA values are systematically lower. AHS and OWA provide stronger and more stable volumetric alignment, although their relative dominance depends on the dataset, partition scheme, and forecasting horizon. In this sense, AHS shows the strongest descriptive robustness across the empirical configurations, particularly in the structurally heterogeneous M5 dataset; however, its advantage over OWA is not uniformly statistically significant after Holm correction. Therefore, the results should be interpreted as evidence of greater descriptive robustness for AHS rather than unconditional statistical superiority over OWA in every configuration.

### 4.1. Walmart

For Walmart, the GRA heatmap in Figure 2 shows that all selectors achieve relatively high median GRA values across horizons and partition schemes. AHS and OWA are consistently competitive, whereas ERA tends to occupy the weakest position. Under the 91:9 partition, AHS is the recommended selector, with 9 wins out of 12 horizons, while OWA leads in the first three horizons. Under the 80:20 and 70:30 partitions, OWA becomes the leading descriptive selector, with 11 and 9 wins, respectively.

However, the inferential evidence indicates that the difference between AHS and OWA should be interpreted cautiously. In all three partitions, the paired Wilcoxon-Holm comparisons show no significant differences between AHS and OWA across the twelve horizons. Thus, Walmart supports OWA as the strongest descriptive selector in the 80:20 and 70:30 partitions, but not as an inferentially separated alternative from AHS.

The selection-frequency heatmap in Figure 3 shows that ERA is concentrated mainly on TRF, whereas AHS and OWA distribute their selections across several models, including KNN, XGB, HR, SES, DTR, and GBR. This indicates that AHS and OWA generate more diversified model assignments across horizons and partition schemes.

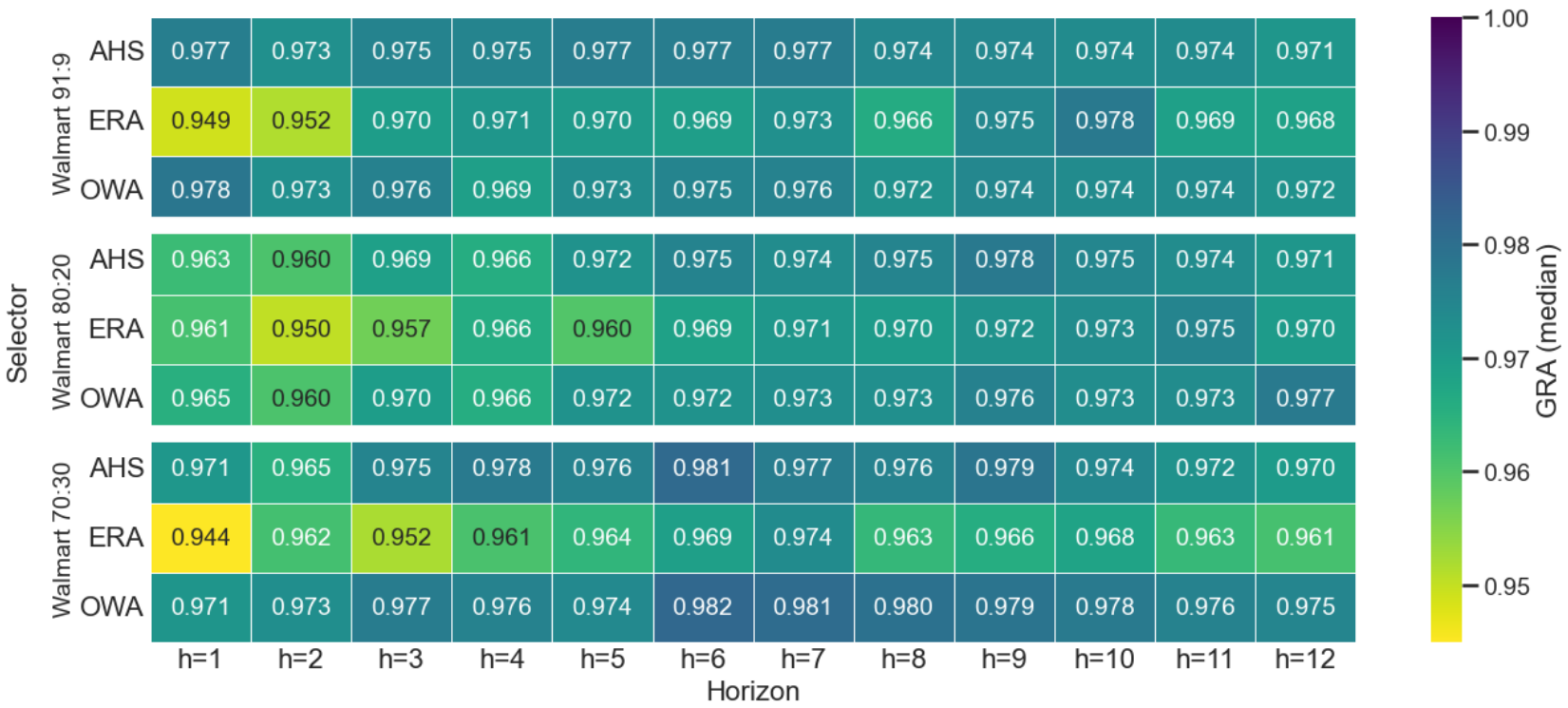

**Figure 2**. GRA heatmap by selector and horizon, Walmart.

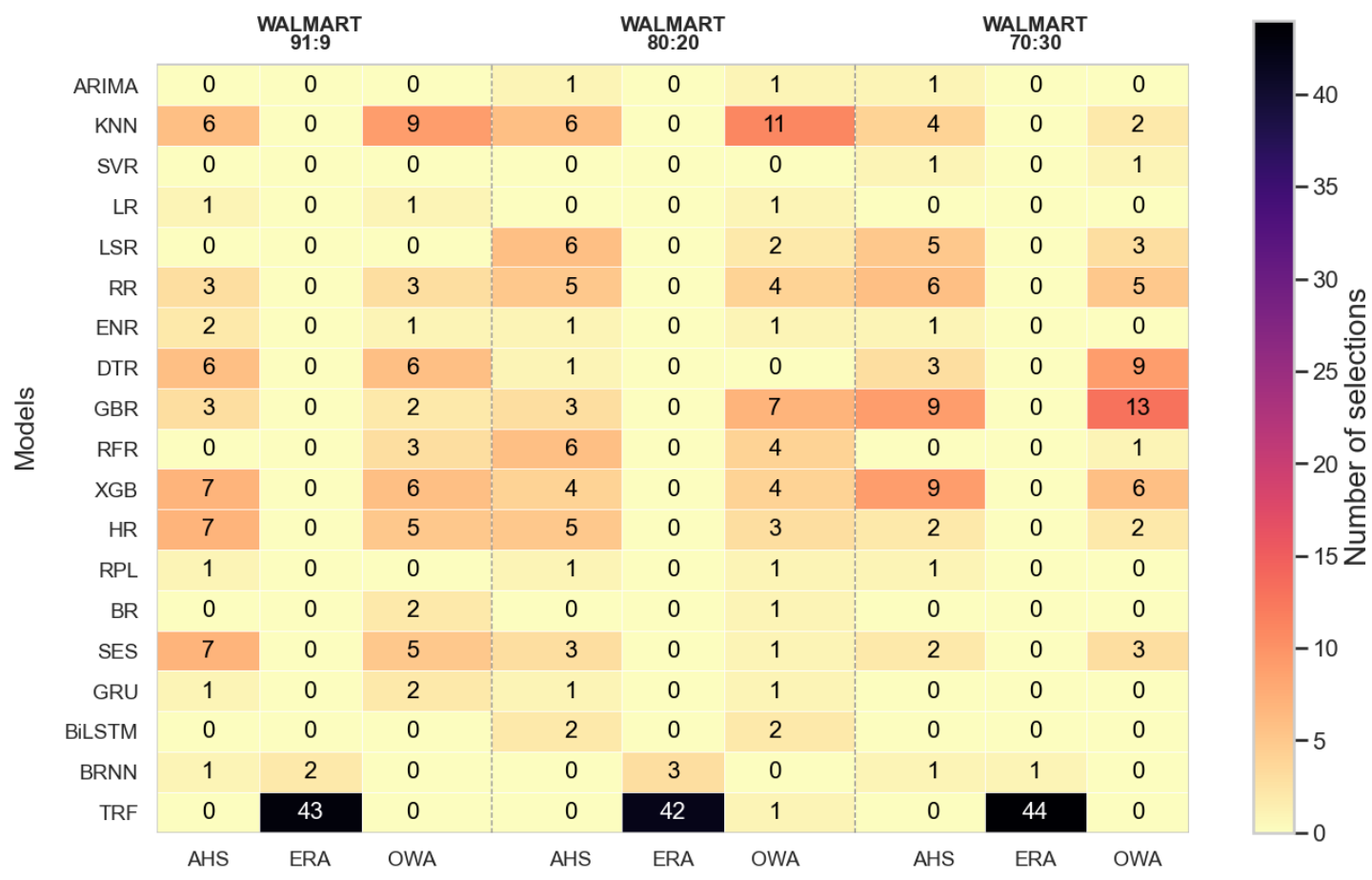

**Figure 3**. Frequency of selected forecasting models by selector and split for Walmart.

### 4.2. M5

For M5, the GRA heatmap in Figure 4 provides the clearest evidence in favor of AHS. Across the three partition schemes, AHS is the recommended selector in all cases, winning 11 horizons under 91:9, all 12 horizons under 80:20, and 11 horizons under 70:30. OWA remains the second-best selector, while ERA exhibits markedly poor volumetric alignment, with strongly negative median GRA values across several horizons.

The inferential results reinforce this pattern. Friedman tests detect significant selector differences in all horizons for the three partitions. In addition, AHS differs significantly from OWA in 11 of 12 horizons under 91:9, in all 12 horizons under 80:20, and in 11 of 12 horizons under 70:30. Therefore, unlike Walmart, the superiority of AHS in M5 is not only descriptive but also strongly supported by the post-hoc comparisons.

The selection-frequency heatmap in Figure 5 shows that ERA is almost entirely concentrated on TRF, whereas AHS and OWA select a broader set of models. AHS assigns models such as KNN, DTR, SES, MLP, CatBoost, PLR, and other alternatives depending on the split. This reinforces the interpretation that AHS is more suitable for highly heterogeneous and sparse demand conditions.

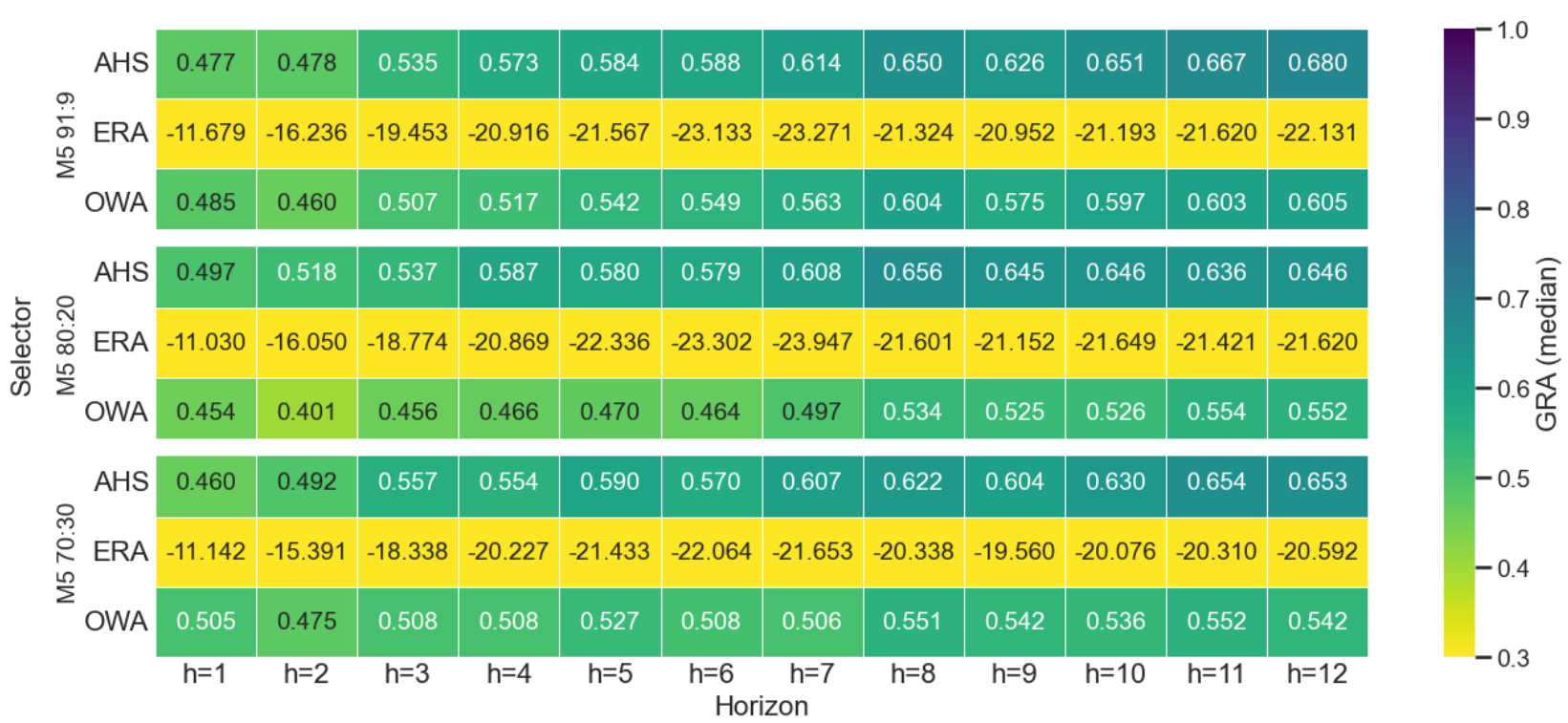

**Figure 4**. GRA heatmap by selector and horizon, M5.

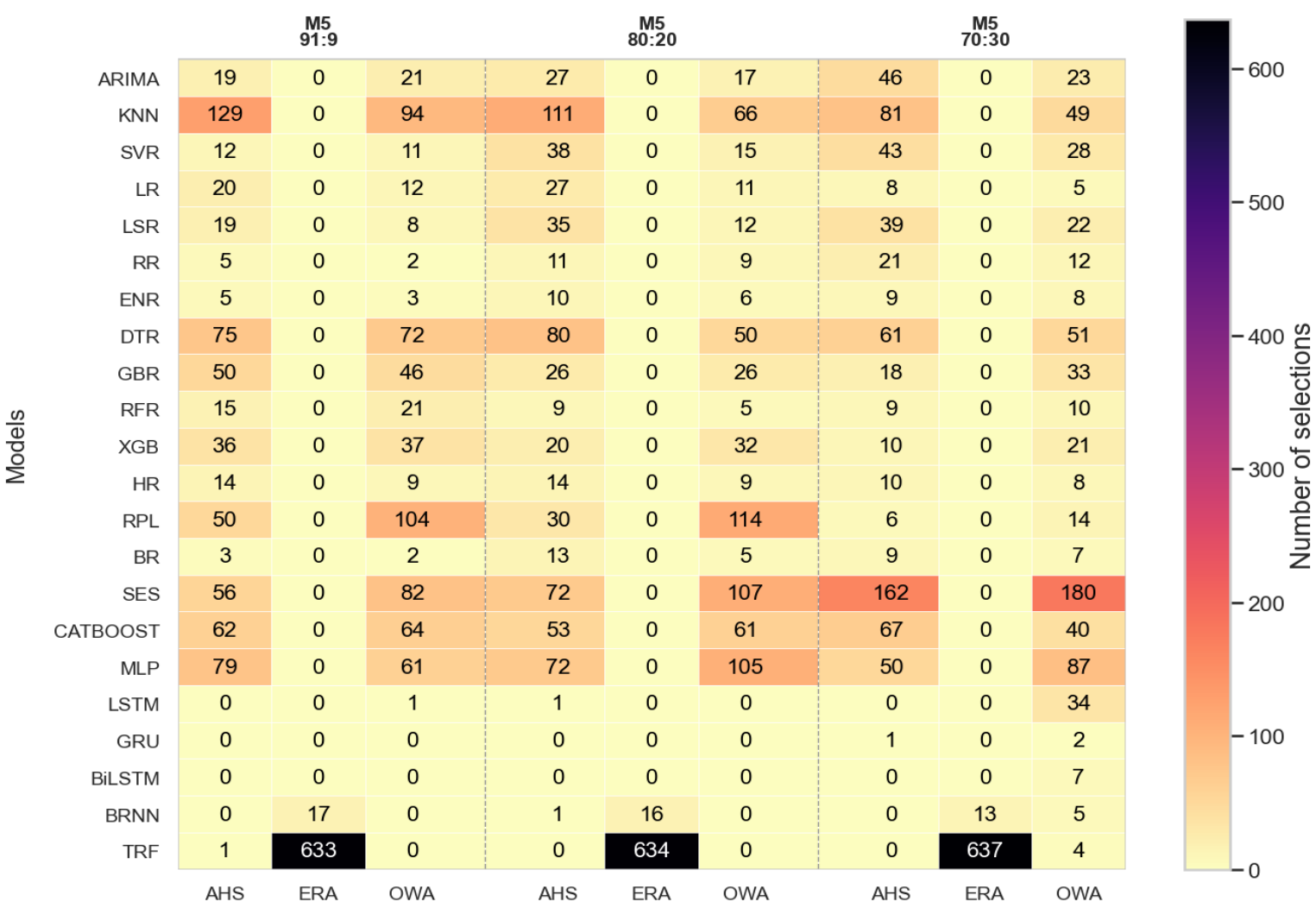

**Figure 5**. Frequency of selected forecasting models by selector and split for M5.

### 4.3. M4

For M4, the GRA heatmap in Figure 6 shows that AHS is the recommended selector in all three partitions, winning all 12 horizons under 91:9, 80:20, and 70:30. AHS and OWA remain very close across horizons, whereas ERA is consistently weaker. This pattern is especially clear in the 91:9 partition, where ERA presents negative median GRA values, while AHS and OWA remain close to the upper range of volumetric alignment.

The inferential analysis confirms that Friedman tests are significant in all horizons across the three partitions. However, the post-hoc results show that these differences are driven mainly by the weaker performance of ERA. AHS and OWA do not differ significantly in any horizon after Holm correction. Therefore, AHS should be interpreted as the preferred descriptive selector for M4, while OWA remains a statistically close alternative.

The selection-frequency heatmap in Figure 7 shows that ERA is again concentrated on TRF, while AHS and OWA assign models more broadly. SES, PLR, GBR, KNN, LR, RR, DTR, and CatBoost appear among the frequently selected alternatives, depending on the split. This indicates that the leading selectors adapt their model assignments more flexibly than ERA.

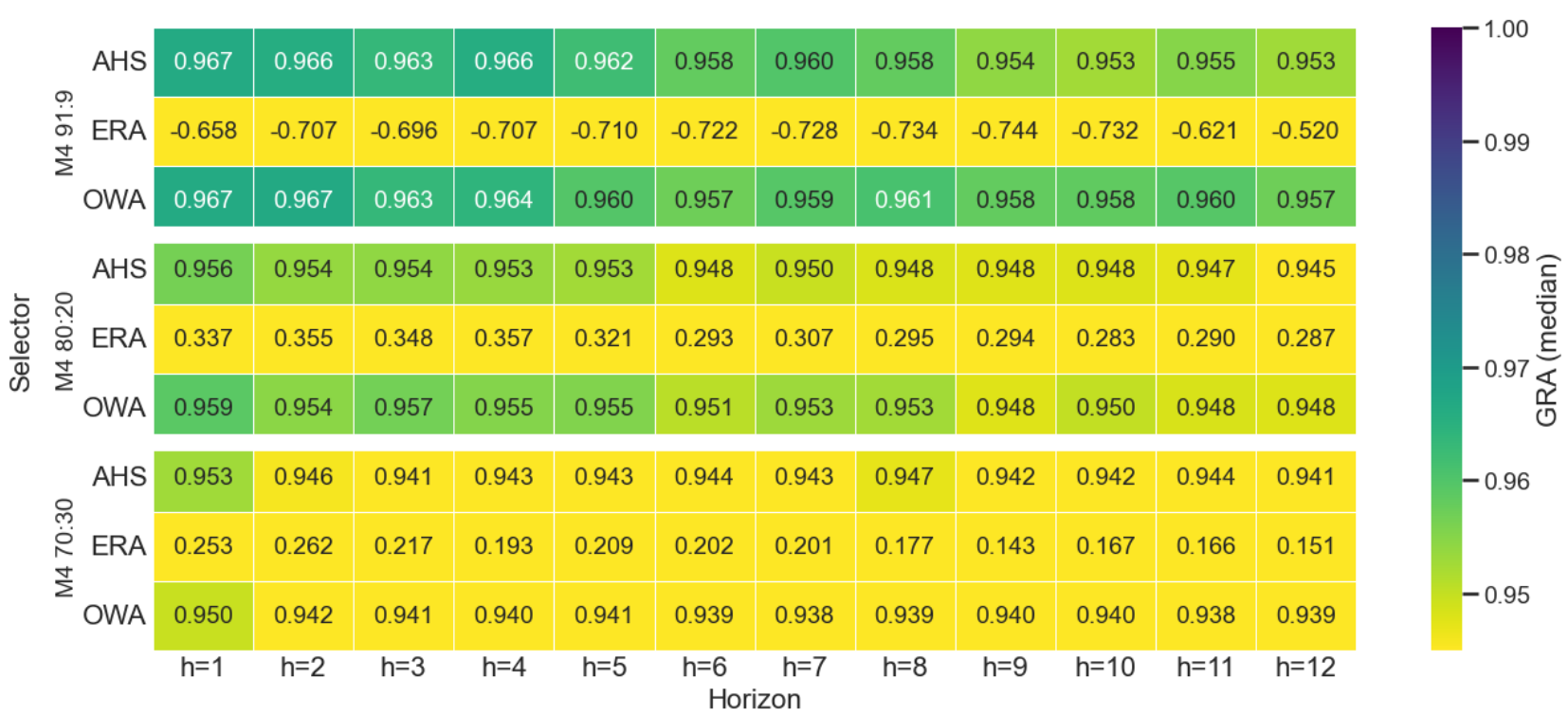

**Figure 6**. GRA heatmap by selector and horizon, M4.

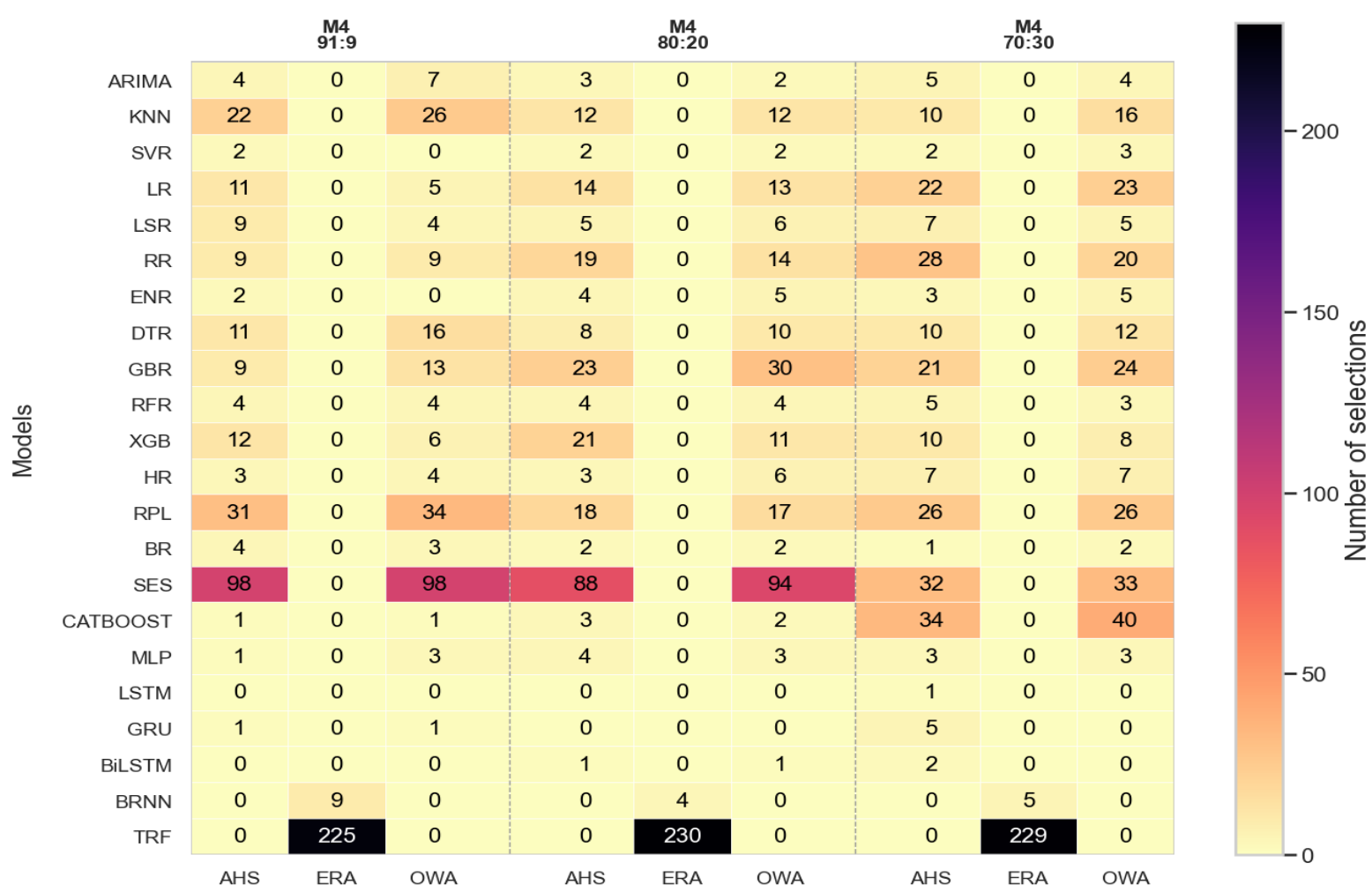

**Figure 7**. Frequency of selected forecasting models by selector and split for M4.

### 4.4. M3

For M3, the GRA heatmap in Figure 8 shows that the relative behavior of AHS and OWA varies across partition schemes. Under 91:9, AHS is the recommended selector, with 8 wins out of 12 horizons, while OWA leads in 4 horizons. Under 80:20, OWA becomes the dominant descriptive selector, winning all 12 horizons. Under 70:30, AHS becomes dominant, also winning all 12 horizons.

The inferential results show significant Friedman tests across all horizons and partitions, indicating systematic differences among selectors. However, as in M4, the strongest separation is mainly associated with the weaker performance of ERA. Under 91:9, AHS and OWA do not differ significantly in any horizon. Under 80:20, OWA differs significantly from AHS in only 2 horizons, and under 70:30, AHS differs significantly from OWA in 4 horizons. Thus, M3 shows that AHS and OWA remain close competitors, although AHS becomes more clearly favorable when the testing proportion increases to 30%.

The selection-frequency heatmap in Figure 9 shows that ERA is strongly concentrated on TRF, whereas AHS and OWA distribute selections across a wider set of models. KNN, SES, DTR, RR, PLR, XGB, LSR, and other models appear repeatedly among the selected alternatives. This supports the view that AHS and OWA provide more flexible assignments across horizons and partition schemes.

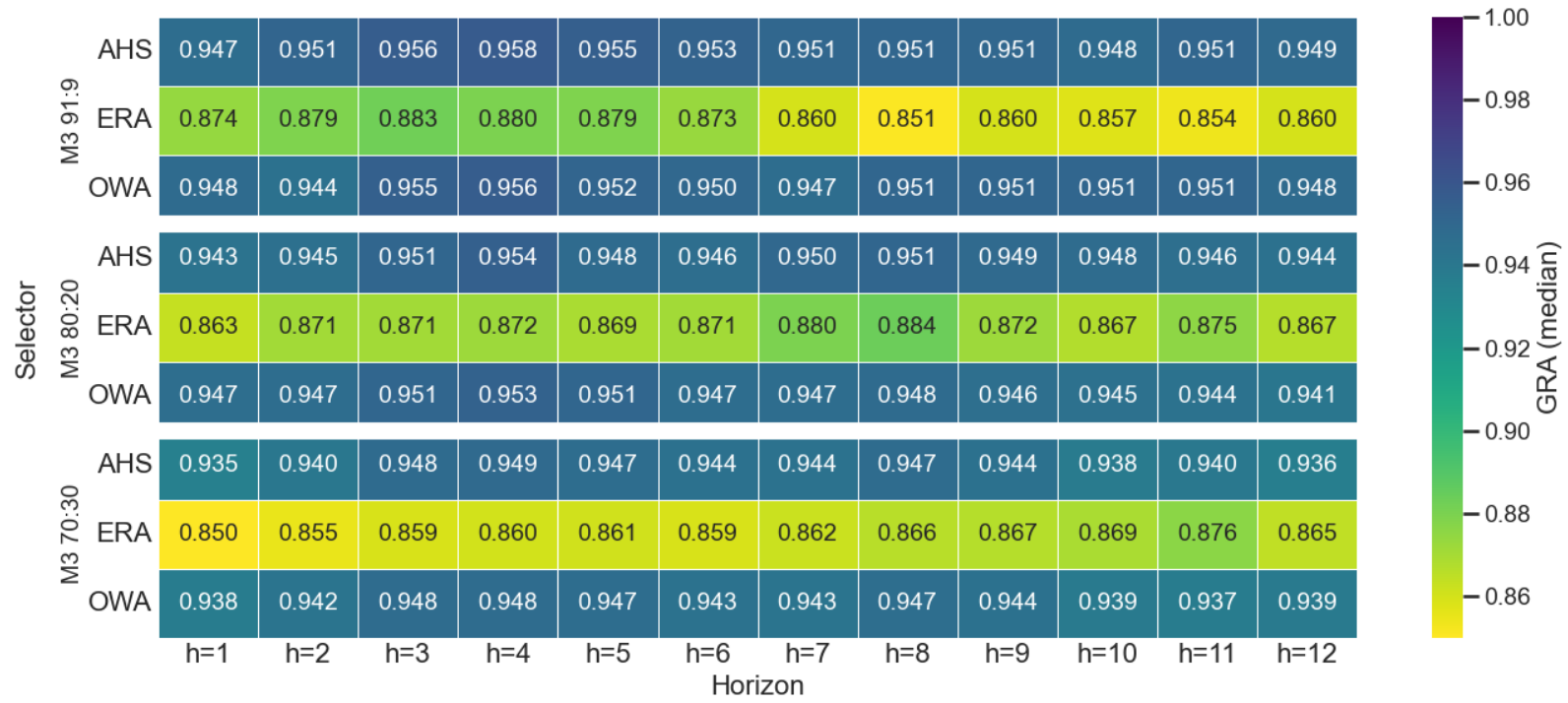

**Figure 8**. GRA heatmap by selector and horizon, M3.

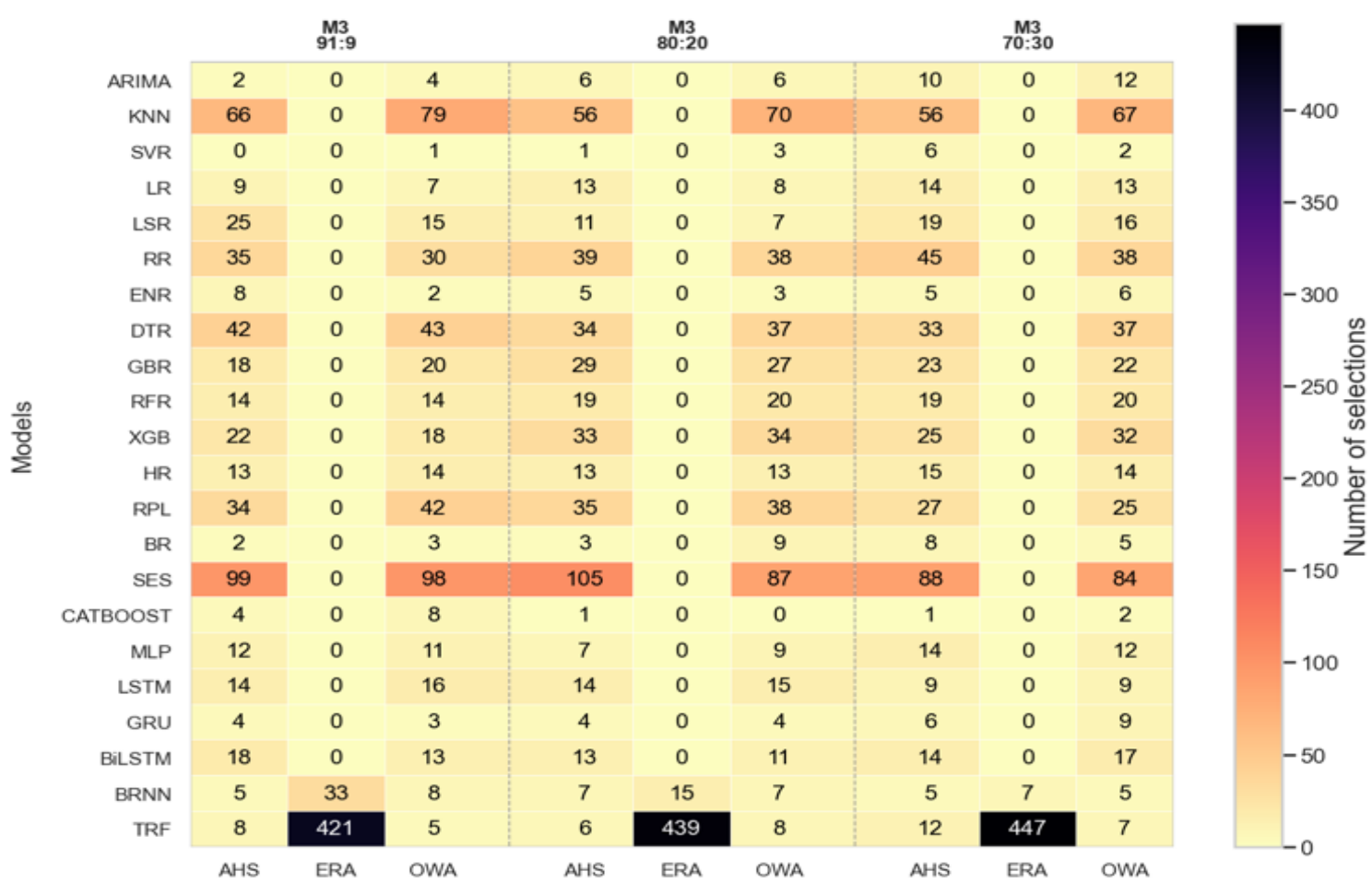

**Figure 9**. Frequency of selected forecasting models by selector and split for M3.

## 4.5. Overall synthesis

Across datasets, AHS is the most robust global selector. It dominates M5 and M4 across all partition schemes and becomes especially strong in M3 under the 70:30 partition. OWA is highly competitive and becomes the preferred descriptive selector in Walmart under 80:20 and 70:30, and in M3 under 80:20. However, in several cases its advantage over AHS is descriptive rather than statistically significant. ERA is systematically the weakest selector, with lower GRA values and a strong tendency to select TRF across datasets and partitions.

Taken together, the results indicate that selector performance is dataset-dependent and horizon-sensitive. AHS provides the strongest overall balance between volumetric coherence, robustness, and adaptive model assignment, while OWA remains a competitive benchmark-relative alternative. ERA is useful as a ranking-aggregation comparator, but its model assignments generally produce weaker ex post volumetric alignment under heterogeneous demand conditions.

# 5. Discusión

The results show that automatic forecasting model selection depends on the dataset structure, the training–testing partition, and the evaluation horizon. The comparison among AHS, OWA, and ERA indicates that no selector is universally dominant across all scenarios. However, two consistent patterns emerge. First, AHS and OWA provide stronger ex post volumetric coherence than ERA. Second, the relative advantage between AHS and OWA varies with the demand structure and the availability of historical data, confirming that model selection should be interpreted as a context-dependent decision problem rather than a universal ranking rule.

## 5.1. Selector performance

AHS shows the most robust overall behavior. Its advantage is especially clear in M5, where it dominates most horizons and presents statistically significant differences against OWA in most partition schemes. This result is consistent with the design of AHS, since the selector incorporates demand-structure information and applies a hierarchical rule that combines scaled, absolute, relative, and bias errors.

OWA also presents competitive performance. It is the descriptively dominant selector in Walmart under the 80:20 and 70:30 partitions, and in M3 under the 80:20 partition. However, in several cases, the differences between OWA and AHS are not statistically significant after Holm correction. Therefore, OWA should be interpreted as a strong benchmark-relative alternative, not as a selector consistently superior to AHS.

ERA performs worst in most configurations. Although it integrates MAE, RMSE, and $R^2$ through ordinal rank aggregation, its model assignments generate lower GRA values. This suggests that multicriteria aggregation based only on ordinal positions does not necessarily produce better ex post volumetric coherence. In addition, the selection-frequency heatmaps show that ERA tends to concentrate its selections on fewer models, especially TRF, thereby reducing its ability to adapt to different demand regimes.

### 5.2. Relationship between demand structure and GRA

The dataset structure, reported in Table 2, helps interpret the results. Walmart is composed exclusively of smooth series, which explains why AHS and OWA obtain high and very similar GRA values. In this context, demand is more regular, and the operational difference between the leading selectors tends to be smaller.

M5 represents the most demanding case, since it combines a high proportion of intermittent and lumpy series with a lower presence of smooth series. In this setting, AHS obtains its clearest advantage. This suggests that explicitly incorporating demand-structure information is especially useful when the series exhibit higher irregularity, dispersion, and forecasting difficulty.

M4 and M3 are mostly composed of smooth series, with a low presence of erratic series. In both cases, AHS usually leads descriptively, but OWA remains statistically close in several horizons and partitions. This indicates that, when demand is structurally more regular, an adaptive rule and a benchmark-relative rule can produce similar volumetric outcomes.

Overall, these results show that the Classes column in Table 2 is not merely descriptive; it helps contextualize the statistical evidence based on GRA. The selectors behave differently depending on the proportion of smooth, intermittent, erratic, and lumpy series, reinforcing the need to consider demand structure in automatic model selection.

### 5.3. Operational interpretation of GRA

GRA evaluates ex post whether the total forecasted volume produced by the selected model remains aligned with the observed demand volume. This perspective is relevant because inventory, purchasing, replenishment, and planning decisions depend not only on pointwise forecast errors but also on the accumulated demand volume expected over several future cycles.

The results show that two selectors may be statistically close while still producing descriptive differences in volumetric alignment. Therefore, GRA does not replace traditional forecast accuracy metrics, nor does it define the forecast's statistical objective. Its role is to evaluate the operational consequences of model selection after the forecasting process.

This interpretation is consistent with recent studies on automatic selection of forecasting models in supply chains. In particular, Garred et al. (2026) use OWA as a selection criterion and discuss automatic selection from a Forecast Value Added perspective, emphasizing that improvements in accuracy should be evaluated together with their usefulness for planning, computational cost, and implementation burden. In line with this perspective, the GRA results connect model selection with an intermediate operational measure of volumetric coherence.

From an inventory perspective, aggregate overestimation may lead to overstock and capital immobilization, whereas persistent underestimation may increase the risk of stockouts and service deterioration. In this sense, GRA acts as an intermediate indicator between statistical forecast evaluation and potential operational impact, although it does not replace an explicit inventory-cost simulation.

### 5.4. Implications for automatic model selection

The selection-frequency results show that AHS and OWA assign models more diversely than ERA. This supports the idea that automatic model selection should not rely on a single globally preferred forecasting model. Instead, model assignment should adapt to the dataset, the forecasting horizon, and the demand structure.

From a computational perspective, the comparison among selectors should be distinguished from the cost of model training and optimization. In this study, the selectors are applied after forecasts have been generated. Therefore, their marginal cost is concentrated in the decision stage. OWA has the lowest computational cost because it only computes benchmark-relative ratios and selects the model with the lowest aggregate score. ERA requires ranking candidate models by MAE, RMSE, and $R^2$, so its cost is higher than OWA but remains moderate. AHS has the highest theoretical cost due to the construction of the Pareto-efficient candidate set. However, in the present experimental design, with (K=22) candidate models, this cost remains limited compared with model training, hyperparameter optimization, and forecast generation.

This interpretation is consistent with the Forecast Value Added logic discussed by Garred et al. (2026), in which accuracy improvements should be evaluated alongside computational burden and practical usefulness for planning. From this perspective, AHS can be justified when its higher decision-layer complexity yields clear gains in volumetric coherence, as observed, especially in M5. Conversely, in more regular datasets such as Walmart, OWA may represent a more parsimonious alternative when its results are statistically close to those of AHS.

Therefore, AHS can be interpreted as an adaptive decision-support layer applied after model training and forecast generation. OWA offers a simpler, interpretable, benchmark-relative alternative. ERA functions as a multicriteria comparator, though its results indicate a weaker ability to achieve volumetric coherence in heterogeneous demand environments.

### 5.5. Limitations and future research

The first limitation is that the validation is based on fixed-origin backtesting. Although this strategy preserves temporal order and avoids information leakage, future studies should evaluate rolling-origin or repeated-window schemes to assess selector stability across different forecast origins.

The second limitation is that GRA measures aggregate volumetric coherence but does not explicitly incorporate inventory costs. Future research should integrate the selected forecasts into inventory-control or production-planning simulations, accounting for holding costs, shortage costs, service levels, replenishment constraints, and production-capacity constraints.

The third limitation is that AHS incorporates a structural condition based on demand frequency and series variability. Therefore, future research should evaluate the sensitivity of this adaptive rule to different values of $p^* and\ c^*$, as well as its stability across other datasets, industries, aggregation levels, and temporal frequencies.

Finally, future research should evaluate the computational cost of the selectors separately from the cost of model training and optimization. This distinction is important because AHS, OWA, and ERA are applied after predictions have been generated. Their marginal cost may be low in configurations with few candidate models, but it may become more relevant in scenarios involving thousands of SKUs, many candidate models, or frequent re-selection processes. Such an analysis would help determine whether improvements in GRA justify the additional complexity of each selection mechanism.

## 6. Conclusion

This study evaluated automatic demand forecasting model selection through three selectors: AHS, OWA, and ERA. The comparison was conducted on the Walmart, M3, M4, and M5 datasets, considering three training–testing partitions and evaluation horizons of up to 12 cycles. Selector performance was assessed using GRA as an ex post indicator of volumetric coherence between the selected models' forecasts and observed demand.

The results show that AHS provides the most robust overall behavior, especially in more heterogeneous contexts such as M5, where the presence of intermittent and lumpy series makes the incorporation of demand-structure information particularly relevant. OWA also shows competitive performance, especially in more regular datasets such as Walmart and in some M3 partitions. In contrast, ERA performs worst across most configurations, suggesting that ordinal aggregation of metrics does not necessarily translate into stronger ex post volumetric coherence.

The evidence also confirms that the demand structure is relevant for interpreting selector performance. The information reported in Table 2, particularly the Classes column, helps explain why AHS obtains clearer advantages in more irregular demand environments, whereas AHS and OWA tend to behave more similarly in datasets dominated by smooth series. Therefore, automatic model selection should not be understood as a universal rule, but rather as a decision process conditioned by the demand regime, the forecasting horizon, and the availability of historical data.

From an operational perspective, GRA complements traditional error metrics by evaluating ex post whether the forecasted volume accumulated by the selected model remains aligned with observed demand over the evaluation horizon. In this sense, it enables analysis of the effect of model selection beyond a single cycle, considering the volumetric coherence of forecasts across several future periods. Although GRA does not replace an explicit inventory-cost simulation, it provides an intermediate measure that links model selection with planning, purchasing, replenishment, and inventory decisions.

Finally, AHS can be interpreted as an adaptive decision-support layer applied after model training and forecast generation. Its higher theoretical complexity relative to OWA may be justified when it produces clear gains in volumetric coherence, as observed in more heterogeneous demand settings. Future research should evaluate the stability of AHS under rolling-origin schemes, analyze the sensitivity of its structural condition based on $p^* and\ c^*$, and integrate the selected forecasts into inventory or production-planning simulations to estimate direct economic effects.

**Author contributions statement**

Adolfo González: Conceptualization, methodology, software, validation, formal analysis, investigation, data curation, writing – original draft, and visualization. Víctor Parada: Supervision and manuscript review. Both authors reviewed and approved the final version of the manuscript.

**Disclosure statement**

The authors declare that they have no known competing financial interests or personal relationships that could have appeared to influence the work reported in this paper.

**Funding**

This research received no external funding.

**Data availability statement**

The datasets analyzed in this study are publicly available. The Walmart source is available via Kaggle (Yasser, 2021): https://www.kaggle.com/code/yasserh/walmart-sales-prediction-best-ml-algorithms. The M3 (Makridakis & Hibon, 2000) data are available from the International Institute of Forecasters repository: https://forecasters.org/resources/time-series-data/m3-competition/. The M4 (Makridakis et al., 2018) data are available from the M4 Competition repository: https://www.m4.unic.ac.cy/the-dataset/. The M5 data are available from the Kaggle M5 (Makridakis et al., 2022) Forecasting Accuracy competition: https://www.kaggle.com/c/m5-forecasting-accuracy.

## Appendix A

**Table A1.** Global selector performance by dataset — Training 91% and Testing 9%.

| Dataset | Recommended selector | AHS wins | OWA wins | ERA wins | Mean position AHS | Mean position OWA | Mean position ERA |
|---|---|---|---|---|---|---|---|
| Walmart | AHS | 9 | 3 | 0 | 1.25 | 1.75 | 3.00 |
| M5 | AHS | 11 | 1 | 0 | 1.08 | 1.92 | 3.00 |
| M4 | AHS | 12 | 0 | 0 | 1.00 | 2.00 | 3.00 |
| M3 | AHS | 8 | 4 | 0 | 1.33 | 1.67 | 3.00 |

**Table A2.** Friedman test summary by dataset — Training 91% and Testing 9%.

| Dataset | Valid horizons | Significant horizons | Non-significant horizons | Main inferential interpretation |
|---|---|---|---|---|
| Walmart | 12 | 4 | 8 | Significant selector differences are concentrated mainly in the shortest horizons, specifically (h = 1), (h = 2), (h = 3), and (h = 5). From (h = 6) onward, no statistically significant global differences are detected, indicating that selector separation becomes weaker in longer horizons. |
| M5 | 12 | 12 | 0 | Strong and systematic selector differences are observed across all forecasting horizons. The inferential separation is mainly driven by the consistently weaker performance of ERA relative to AHS and OWA. In addition, AHS differs significantly from OWA in 11 of the 12 horizons, indicating a more clearly separated leading behavior for AHS under the M5 split. |
| M4 | 12 | 12 | 0 | Strong and systematic selector differences are observed across all forecasting horizons. However, these differences are mainly driven by the weaker behavior of ERA. AHS and OWA remain statistically close across all horizons, so the superiority of AHS over OWA should be interpreted mainly as descriptive rather than inferential. |
| M3 | 12 | 12 | 0 | Strong and systematic selector differences are observed across all forecasting horizons. However, as in M4, the inferential separation is mainly explained by the weaker behavior of ERA. AHS and OWA remain statistically similar across all horizons, although AHS obtains the best descriptive global performance. |

**Table A3.** Post-hoc Wilcoxon-Holm summary — Training 91% and Testing 9%.

| Dataset | ERA vs OWA significant cases | ERA vs AHS significant cases | OWA vs AHS significant cases | Interpretation |
|---|---|---|---|---|
| Walmart | 3/12 | 4/12 | 0/12 | The statistically significant post-hoc differences are limited and mainly associated with the separation of ERA from AHS and OWA in the shortest horizons. AHS and OWA remain statistically similar across all twelve horizons, indicating strong proximity between the two leading selectors. |
| M5 | 12/12 | 12/12 | 11/12 | ERA differs systematically from both AHS and OWA across all horizons. Unlike Walmart, AHS and OWA also differ significantly in almost all horizons, except (h = 1), where the difference is only descriptive. This indicates a stronger inferential advantage of AHS in M5. |
| M4 | 12/12 | 12/12 | 0/12 | ERA is consistently separated from both AHS and OWA across all horizons. No significant differences are detected between AHS and OWA, indicating strong statistical proximity between the two leading selectors despite the systematic descriptive advantage of AHS. |
| M3 | 12/12 | 12/12 | 0/12 | ERA differs significantly from both AHS and OWA in every forecasting horizon. However, AHS and OWA do not differ significantly in any horizon after Holm correction, indicating that their difference is descriptive rather than inferential. |

**Table A4.** Descriptive GRA summary by dataset — Training 91% and Testing 9%.

| Dataset | Best selector by GRA global | Second selector | Weakest selector | Main descriptive finding |
|---|---|---|---|---|
| Walmart | AHS | OWA | ERA | AHS leads most horizons, with 9 wins out of 12, especially from (h = 4) to (h = 12). OWA leads in the first three horizons, while ERA does not lead in any horizon and remains the weakest selector in the descriptive GRA-based ranking. |
| M5 | AHS | OWA | ERA | AHS shows a dominant descriptive pattern, leading in 11 of the 12 horizons. OWA leads only at (h = 1), and its advantage at that horizon is not statistically significant after Holm correction. ERA remains systematically last and exhibits markedly weaker GRA behavior across the full forecasting horizon. |
| M4 | AHS | OWA | ERA | AHS leads all twelve horizons according to GRA global, with a perfect descriptive dominance pattern. However, AHS and OWA remain statistically similar in all paired Holm-corrected comparisons. ERA remains systematically last and is significantly worse than both leading selectors across all horizons. |
| M3 | AHS | OWA | ERA | AHS leads 8 of the 12 horizons, especially from (h = 6) to (h = 12), while OWA leads in 4 horizons, mainly at shorter horizons. ERA does not lead in any horizon and is consistently weaker than both leading selectors. |

**Table A5.** Global selector performance by dataset — Training 80% and Testing 20%.

| Dataset | Recommended selector | AHS wins | OWA wins | ERA wins | Mean position AHS | Mean position OWA | Mean position ERA |
|---|---|---|---|---|---|---|---|
| Walmart | OWA | 1 | 11 | 0 | 1.92 | 1.08 | 3.00 |
| M5 | AHS | 12 | 0 | 0 | 1.00 | 2.00 | 3.00 |
| M4 | AHS | 12 | 0 | 0 | 1.00 | 2.00 | 3.00 |
| M3 | OWA | 0 | 12 | 0 | 2.00 | 1.00 | 3.00 |

**Table A6.** Friedman test summary by dataset — Training 80% and Testing 20%.

| Dataset | Valid horizons | Significant horizons | Non-significant horizons | Main inferential interpretation |
|---|---|---|---|---|
| Walmart | 12 | 1 | 11 | Global selector differences are weak under the 80:20 split. Friedman detects significant differences in only one horizon, so the dominance of OWA should be interpreted mainly as descriptive rather than inferential. |
| M5 | 12 | 12 | 0 | Strong and systematic selector differences are observed across all forecasting horizons. AHS dominates all horizons and is significantly separated from OWA and ERA across the full horizon range. |
| M4 | 12 | 12 | 0 | Strong and systematic selector differences are observed across all forecasting horizons. However, these differences are mainly driven by the weaker behavior of ERA, since AHS and OWA remain statistically close in all horizons. |
| M3 | 12 | 12 | 0 | Strong and systematic selector differences are observed across all forecasting horizons. OWA dominates all horizons descriptively, while ERA remains consistently weaker. AHS and OWA differ significantly only in a limited number of horizons. |

**Table A7.** Post-hoc Wilcoxon-Holm summary — Training 80% and Testing 20%.

| Dataset | ERA vs OWA significant cases | ERA vs AHS significant cases | OWA vs AHS significant cases | Interpretation |
|---|---|---|---|---|
| Walmart | 2/12 | 1/12 | 0/12 | Post-hoc differences are limited. OWA leads descriptively in 11 horizons, but it does not differ significantly from AHS in any horizon after Holm correction. |
| M5 | 12/12 | 12/12 | 12/12 | AHS is consistently and significantly separated from both OWA and ERA across all horizons. This provides strong inferential support for AHS under the 80:20 split. |
| M4 | 12/12 | 12/12 | 0/12 | ERA is consistently separated from both AHS and OWA. However, AHS and OWA do not differ significantly in any horizon, indicating strong statistical proximity between the two leading selectors. |
| M3 | 12/12 | 12/12 | 2/12 | ERA differs significantly from both AHS and OWA in all horizons. OWA is the dominant descriptive selector, but its separation from AHS is statistically significant only in two horizons. |

**Table A8.** Descriptive GRA summary by dataset — Training 80% and Testing 20%.

| Dataset | Best selector by GRA_global | Second selector | Weakest selector | Main descriptive finding |
|---|---|---|---|---|
| **Walmart** | OWA | AHS | ERA | OWA leads 11 of the 12 horizons, showing strong descriptive dominance under the 80:20 split. AHS leads only one horizon, while ERA remains the weakest selector. However, OWA and AHS are not significantly different after Holm correction. |
| **M5** | AHS | OWA | ERA | AHS leads all twelve horizons and is statistically superior to OWA in every horizon. ERA remains systematically last and shows markedly weaker GRA behavior. |
| **M4** | AHS | OWA | ERA | AHS leads all twelve horizons according to GRA_global. Nevertheless, its advantage over OWA is descriptive, because no significant post-hoc differences are detected between AHS and OWA. ERA remains consistently weaker than both. |
| **M3** | OWA | AHS | ERA | OWA leads all twelve horizons according to GRA_global, indicating complete descriptive dominance. However, AHS remains close, with significant OWA–AHS differences detected only in two horizons. ERA does not lead in any horizon and remains the weakest selector. |

**Table A9.** Global selector performance by dataset — Training 70% and Testing 30%.

| Dataset | Recommended selector | AHS wins | OWA wins | ERA wins | Mean position AHS | Mean position OWA | Mean position ERA |
|---|---|---|---|---|---|---|---|
| **Walmart** | OWA | 3 | 9 | 0 | 1.75 | 1.25 | 3.00 |
| **M5** | AHS | 11 | 1 | 0 | 1.08 | 1.92 | 3.00 |
| **M4** | AHS | 12 | 0 | 0 | 1.00 | 2.00 | 3.00 |
| **M3** | AHS | 12 | 0 | 0 | 1.00 | 2.00 | 3.00 |

**Table A10.** Friedman test summary by dataset — Training 70% and Testing 30%.

| Dataset | Valid horizons | Significant horizons | Non-significant horizons | Main inferential interpretation |
|---|---|---|---|---|
| **Walmart** | 12 | 9 | 3 | Global selector differences are detected in most horizons. OWA leads descriptively in 9 of 12 horizons, but the paired post-hoc comparison between |
| | | | | OWA and AHS is not significant in any horizon. Therefore, the advantage of OWA over AHS should be interpreted mainly as descriptive. |
| **M5** | 12 | 12 | 0 | Strong and systematic selector differences are observed across all forecasting horizons. AHS dominates 11 of 12 horizons, while OWA leads only at (h = 1). The AHS–OWA separation is significant in most horizons, indicating a strong inferential advantage of AHS after the first horizon. |
| **M4** | 12 | 12 | 0 | Strong and systematic selector differences are observed across all horizons. These differences are driven primarily by the weaker behavior of ERA, because AHS and OWA remain statistically close across all horizons. AHS is therefore the preferred descriptive selector, while OWA remains a statistically close alternative. |
| **M3** | 12 | 12 | 0 | Strong and systematic selector differences are detected across all horizons. AHS leads all twelve horizons and differs from OWA in a subset of horizons, indicating stronger evidence for AHS than in the 91:9 and 80:20 splits. ERA remains consistently weaker than both leading selectors. |

**Table A11.** Post-hoc Wilcoxon-Holm summary — Training 70% and Testing 30%.

| Dataset | ERA vs OWA significant cases | ERA vs AHS significant cases | OWA vs AHS significant cases | Interpretation |
|---|---|---|---|---|
| **Walmart** | 12/12 | 11/12 | 0/12 | ERA is clearly separated from OWA in all horizons and from AHS in almost all horizons. However, OWA and AHS do not differ significantly in any horizon, so the OWA advantage is descriptive rather than inferential. |
| **M5** | 12/12 | 12/12 | 11/12 | ERA differs systematically from both AHS and OWA across all horizons. AHS and OWA differ significantly in 11 of the 12 horizons, supporting the inferential advantage of AHS in the majority of the horizon range. |
| **M4** | 12/12 | 12/12 | 0/12 | ERA is consistently separated from both AHS and OWA. No significant differences are detected between AHS and OWA, indicating strong statistical proximity between the two leading selectors despite the complete descriptive dominance of AHS. |
| **M3** | 12/12 | 12/12 | 4/12 | ERA differs significantly from both AHS and OWA in every forecasting horizon. AHS is the dominant descriptive selector, and its advantage over OWA is statistically significant in four horizons. |

**Table A12.** Descriptive GRA summary by dataset — Training 70% and Testing 30%.

| Dataset | Best selector by GRA global | Second selector | Weakest selector | Main descriptive finding |
|---|---|---|---|---|
| **Walmart** | OWA | AHS | ERA | OWA leads 9 of the 12 horizons, while AHS leads 3 horizons. ERA does not lead in any horizon and remains the weakest selector. However, OWA and AHS are not significantly different after Holm correction in any horizon. |
| **M5** | AHS | OWA | ERA | AHS leads 11 of the 12 horizons, with OWA leading only at (h = 1). The difference between AHS and OWA is statistically significant in 11 horizons, providing strong inferential support for AHS as the global selector. |
| **M4** | AHS | OWA | ERA | AHS leads all twelve horizons according to GRA global. Nevertheless, its advantage over OWA is descriptive because no significant post-hoc differences are detected between AHS and OWA. ERA remains consistently weaker than both. |
| **M3** | AHS | OWA | ERA | AHS leads all twelve horizons, indicating complete descriptive dominance under the 70:30 split. Unlike the 91:9 split, AHS also shows statistically significant separation from OWA in four horizons. ERA remains the weakest selector across the full horizon. |

## References

Abadi, M., Barham, P., Chen, J., Chen, Z., Davis, A., Dean, J., Devin, M., Ghemawat, S., Irving, G., Isard, M., Kudlur, M., Levenberg, J., Monga, R., Moore, S., Murray, D. G., Steiner, B., Tucker, P., Vasudevan, V., Warden, P., … Zheng, X. (2016). *TensorFlow: A System for Large-Scale Machine Learning*. 265-283. https://www.usenix.org/conference/osdi16/technical-sessions/presentation/abadi

Abdallah, M., Rossi, R. A., Mahadik, K., Kim, S., Zhao, H., & Bagchi, S. (2025). Evaluation-free Time-series Forecasting Model Selection via Meta-learning. *ACM Transactions on Knowledge Discovery from Data*, *19*(3), 1-41. https://doi.org/10.1145/3715149

Aidoo-Anderson, A., Polychronakis, Y., Sapountzis, S. (Stelios), & Kelly, S. (2025). Investigating demand forecasting practices and challenges in Ghana's Manufacturing Pharmaceutical (MPharma) small and medium enterprises (SMEs): Insights and recommendations. *International Journal of Production Research*, *63*(21), 7899-7920. https://doi.org/10.1080/00207543.2025.2508335

Akiba, T., Sano, S., Yanase, T., Ohta, T., & Koyama, M. (2019). Optuna: A Next-generation Hyperparameter Optimization Framework. *Proceedings of the 25th ACM SIGKDD International Conference on Knowledge Discovery & Data Mining*, 2623-2631. https://doi.org/10.1145/3292500.3330701

Altman, N. S. (1992). An Introduction to Kernel and Nearest-Neighbor Nonparametric Regression. *The American Statistician*, *46*(3), 175-185. https://doi.org/10.1080/00031305.1992.10475879

Armstrong, J. S. (2001). *Principles of Forecasting: A Handbook for Researchers and Practitioners* (Vol. 30). Springer Science & Business Media.

Badulescu, Y., Hameri, A.-P., & Cheikhrouhou, N. (2021). Evaluating demand forecasting models using multi-criteria decision-making approach. *Journal of Advances in Management Research*, *18*(5), 661-683. https://doi.org/10.1108/JAMR-05-2020-0080

Box, G. E. P., Jenkins, G. M., Reinsel, G. C., & Ljung, G. M. (2015). *Time Series Analysis: Forecasting and Control*. John Wiley & Sons.

Breiman, L. (2001). Random Forests. *Machine Learning*, *45*(1), 5-32. https://doi.org/10.1023/A:1010933404324

Breiman, L., Friedman, J., Olshen, R. A., & Stone, C. J. (2017). *Classification and regression trees*. Chapman and Hall/CRC.

Brown, R. G. (1959). Statistical forecasting for inventory control. *(No Title)*.

Chen, T., & Guestrin, C. (2016). XGBoost: A Scalable Tree Boosting System. *Proceedings of the 22nd ACM SIGKDD International Conference on Knowledge Discovery and Data Mining*, 785-794. https://doi.org/10.1145/2939672.2939785

Cho, K., Van Merrienboer, B., Gulcehre, C., Bahdanau, D., Bougares, F., Schwenk, H., & Bengio, Y. (2014). Learning Phrase Representations using RNN Encoder–Decoder for Statistical Machine Translation. *Proceedings of the 2014 Conference on Empirical Methods in Natural Language Processing (EMNLP)*, 1724-1734. https://doi.org/10.3115/v1/D14-1179

Chollet, F. (2015). *Keras* [Python]. https://github.com/fchollet/keras

Demizu, T., Fukazawa, Y., & Morita, H. (2023). Inventory management of new products in retailers using model-based deep reinforcement learning. *Expert Systems with Applications*, *229*, 120256. https://doi.org/10.1016/j.eswa.2023.120256

Doszyń, M., & Dudek, A. (2024). Outliers in intermittent demand forecasting. *Procedia Computer Science*, *246*, 4114-4122. https://doi.org/10.1016/j.procs.2024.09.250

Erjiang, E., Ming, Y., Xin, T., & Ye, T. (2022). Dynamic Model Selection Based on Demand Pattern Classification in Retail Sales Forecasting. *Mathematics*, *10*(17), 3179. https://doi.org/10.3390/math10173179

Fildes, R., & Petropoulos, F. (2015). Simple versus complex selection rules for forecasting many time series. *Journal of Business Research*, *68*(8), 1692-1701. https://doi.org/10.1016/j.jbusres.2015.03.028

Flyckt, J., & Lavesson, N. (2025). Navigating Demand Forecasting in Make-to-Order Manufacturing: The Role of Global Models and Intermittent Time-Series. *Proceedings of the Swedish AI Society Workshop 2025*, *4037*, 12-25.

Friedman, J. H. (2001). Greedy function approximation: A gradient boosting machine. *The Annals of Statistics*, *29*(5). https://doi.org/10.1214/aos/1013203451

Garred, W., Oger, R., & Lauras, M. (2026). Automatic demand forecast model selection in supply chains: A forecast value-added analysis of selection strategies, machine learning, and hyperparameter optimisation. *International Journal of Production Research*, 1-20. https://doi.org/10.1080/00207543.2026.2623194

Giannopoulos, P. G., Dasaklis, T. K., Tsantilis, I., & Patsakis, C. (2025). Machine learning algorithms in intermittent demand forecasting: A review. *International Journal of Production Research*, 1-43. https://doi.org/10.1080/00207543.2025.2578701

Gneiting, T. (2011). Making and Evaluating Point Forecasts. *Journal of the American Statistical Association*, *106*(494), 746-762. https://doi.org/10.1198/jasa.2011.r10138

Goltsos, T. E., Syntetos, A. A., Glock, C. H., & Ioannou, G. (2022). Inventory – forecasting: Mind the gap. *European Journal of Operational Research*, *299*(2), 397-419. https://doi.org/10.1016/j.ejor.2021.07.040

González, A. (2020). Un modelo de gestión de inventarios basado en estrategia competitiva. *Ingeniare. Revista chilena de ingeniería*, *28*(1), 133-142. https://doi.org/10.4067/S0718-33052020000100133

González, A., & Parada, V. (2026). Hierarchical evaluation function: A multi-metric approach for optimizing demand forecasting models. *Expert Systems with Applications*, *311*, 131289. https://doi.org/10.1016/j.eswa.2026.131289

Goodwin, P., Hoover, J., Makridakis, S., Petropoulos, F., & Tashman, L. (2023). Business forecasting methods: Impressive advances, lagging implementation. *PLOS ONE*, *18*(12), e0295693. https://doi.org/10.1371/journal.pone.0295693

Graves, A., & Schmidhuber, J. (2005). Framewise phoneme classification with bidirectional LSTM and other neural network architectures. *Neural Networks*, *18*(5-6), 602-610. https://doi.org/10.1016/j.neunet.2005.06.042

Harris, C. R., Millman, K. J., Van Der Walt, S. J., Gommers, R., Virtanen, P., Cournapeau, D., Wieser, E., Taylor, J., Berg, S., Smith, N. J., Kern, R., Picus, M., Hoyer, S., Van Kerkwijk, M. H., Brett, M., Haldane, A., Del Río, J. F., Wiebe, M., Peterson, P., … Oliphant, T. E. (2020). Array programming with NumPy. *Nature*, *585*(7825), 357-362. https://doi.org/10.1038/s41586-020-2649-2

Haykin, S. (1994). *Neural networks: A comprehensive foundation*. Prentice hall PTR.

Hendry, D. F., & Pretis, F. (2023). Analysing differences between scenarios. *International Journal of Forecasting*, *39*(2), 754-771. https://doi.org/10.1016/j.ijforecast.2022.02.004

Hochreiter, S., & Schmidhuber, J. (1997). Long Short-Term Memory. *Neural Computation*, *9*(8), 1735-1780. https://doi.org/10.1162/neco.1997.9.8.1735

Hoerl, A. E., & Kennard, R. W. (1970). Ridge Regression: Biased Estimation for Nonorthogonal Problems. *Technometrics*, *12*(1), 55-67. https://doi.org/10.1080/00401706.1970.10488634

Huber, J., Gossmann, A., & Stuckenschmidt, H. (2017). Cluster-based hierarchical demand forecasting for perishable goods. *Expert Systems with Applications*, *76*, 140-151. https://doi.org/10.1016/j.eswa.2017.01.022

Huber, P. J. (1964). Robust Estimation of a Location Parameter. *The Annals of Mathematical Statistics*, *35*(1), 73-101. https://doi.org/10.1214/aoms/1177703732

Hunter, J. D. (2007). Matplotlib: A 2D Graphics Environment. *Computing in Science & Engineering*, *9*(3), 90-95. https://doi.org/10.1109/MCSE.2007.55

Hyndman, R. J., & Athanasopoulos, G. (2021). *Forecasting: Principles and Practice* (3.ª ed.). OTexts. https://otexts.com/fpp3/

Hyndman, R. J., & Koehler, A. B. (2006). Another look at measures of forecast accuracy. *International Journal of Forecasting*, *22*(4), 679-688. https://doi.org/10.1016/j.ijforecast.2006.03.001

Khan, N. T., & Al Hanbali, A. (2025). Machine Learning Approaches for Disaggregated and Intermittent Demand Forecasting for Last-mile Logistics. *Transportation Research Procedia*, *84*, 307-314. https://doi.org/10.1016/j.trpro.2025.03.077

Kolassa, S. (2020). Why the “best” point forecast depends on the error or accuracy measure. *International Journal of Forecasting*, *36*(1), 208-211. https://doi.org/10.1016/j.ijforecast.2019.02.017

Kolassa, S., & Schütz, W. (2007). Advantages of the MAD/Mean Ratio over the MAPE. *Foresight: The International Journal of Applied Forecasting*, (6), 40-43.

Kourentzes, N., Barrow, D. K., & Crone, S. F. (2014). Neural network ensemble operators for time series forecasting. *Expert Systems with Applications*, *41*(9), 4235-4244. https://doi.org/10.1016/j.eswa.2013.12.011

Kumar, V. D., Maheswari, S., Raman, Y. S., Iniyavan, S., & Hashim, S. A. (2025). Inventory Optimization and Demand Forecasting Using Machine Learning. En G. Dilip, P. Durgadevi, K. Akila, S. Sridevi, R. Ramachandran, & J. Tipmontian (Eds.), *Proceedings of the International Conference on Intelligent Systems and Digital Transformation (ICISD 2025)* (Vol. 15, pp. 760-773). Atlantis Press International BV. https://doi.org/10.2991/978-94-6463-866-0_62

MacKay, D. J. C. (1992). Bayesian Interpolation. *Neural Computation*, *4*(3), 415-447. https://doi.org/10.1162/neco.1992.4.3.415

Makridakis, S., & Hibon, M. (2000). The M3-Competition: Results, conclusions and implications. *International Journal of Forecasting*, *16*(4), 451-476. https://doi.org/10.1016/S0169-2070(00)00057-1

Makridakis, S., & Petropoulos, F. (2020). The M4 competition: Conclusions. *International Journal of Forecasting*, *36*(1), 224-227. https://doi.org/10.1016/j.ijforecast.2019.05.006

Makridakis, S., Spiliotis, E., & Assimakopoulos, V. (2018). The M4 Competition: Results, findings, conclusion and way forward. *International Journal of Forecasting*, *34*(4), 802-808. https://doi.org/10.1016/j.ijforecast.2018.06.001

Makridakis, S., Spiliotis, E., & Assimakopoulos, V. (2022). M5 accuracy competition: Results, findings, and conclusions. *International Journal of Forecasting*, *38*(4), 1346-1364. https://doi.org/10.1016/j.ijforecast.2021.11.013

Mentzer, J. T., & Moon, M. A. (2004). *Sales Forecasting Management: A Demand Management Approach*. Sage Publications.

Montgomery, D. C., Peck, E. A., & Vining, G. G. (2021). *Introduction to linear regression analysis*. John Wiley & Sons.

Nguyen, T. N. A. (2026). Optimizing supply chain operations using advanced Time-Series Mixer models for demand forecasting and inventory under uncertain demand. *Expert Systems With Applications*.

Paszke, A., Gross, S., Massa, F., Lerer, A., Bradbury, J., Chanan, G., Killeen, T., Lin, Z., Gimelshein, N., Antiga, L., Desmaison, A., Kopf, A., Yang, E., DeVito, Z., Raison, M., Tejani, A., Chilamkurthy, S., Steiner, B., Fang, L., … Chintala, S. (2019). PyTorch: An Imperative Style, High-Performance Deep Learning Library. En H. Wallach, H. Larochelle, A. Beygelzimer, F. d'Alché-Buc, E. Fox, & R. Garnett (Eds.), *Advances in Neural Information Processing Systems* (Vol. 32). Curran Associates, Inc. https://proceedings.neurips.cc/paper_files/paper/2019/file/bdbca288fee7f92f2bfa9f7012727740-Paper.pdf

Pedregosa, F., Varoquaux, G., Gramfort, A., Michel, V., Thirion, B., Grisel, O., Blondel, M., Prettenhofer, P., Weiss, R., Dubourg, V., Vanderplas, J., Passos, A., Cournapeau, D., Brucher, M., Perrot, M., & Duchesnay, É. (2011). Scikit-learn: Machine Learning in Python. *Journal of Machine Learning Research*, *12*(85), 2825-2830.

Petropoulos, F., Apiletti, D., Assimakopoulos, V., Babai, M. Z., Barrow, D. K., Ben Taieb, S., Bergmeir, C., Bessa, R. J., Bijak, J., Boylan, J. E., Browell, J., Carnevale, C., Castle, J. L., Cirillo, P., Clements, M. P., Cordeiro, C., Cyrino Oliveira, F. L., De Baets, S., Dokumentov, A., … Ziel, F. (2022). Forecasting: Theory and practice. *International Journal of Forecasting*, *38*(3), 705-871. https://doi.org/10.1016/j.ijforecast.2021.11.001

Poler, R., & Mula, J. (2011). Forecasting model selection through out-of-sample rolling horizon weighted errors. *Expert Systems with Applications*, *38*(12), 14778-14785. https://doi.org/10.1016/j.eswa.2011.05.072

Prokhorenkova, L., Gusev, G., Vorobev, A., Dorogush, A. V., & Gulin, A. (2018). CatBoost: Unbiased boosting with categorical features. *Proceedings of the 32nd International Conference on Neural Information Processing Systems, NIPS'18*, 6639-6649.

Sanders, N. R., & Graman, G. A. (2016). Impact of Bias Magnification on Supply Chain Costs: The Mitigating Role of Forecast Sharing. *Decision Sciences*, *47*(5), 881-906. https://doi.org/10.1111/deci.12208

Sayed, H. E., Gabbar, H. A., & Miyazaki, S. (2009). A hybrid statistical genetic-based demand forecasting expert system. *Expert Systems with Applications*, *36*(9), 11662-11670. https://doi.org/10.1016/j.eswa.2009.03.014

Schuster, M., & Paliwal, K. K. (1997). Bidirectional recurrent neural networks. *IEEE Transactions on Signal Processing*, *45*(11), 2673-2681. https://doi.org/10.1109/78.650093

Seabold, S., & Perktold, J. (2010). *Statsmodels: Econometric and Statistical Modeling with Python*. 92-96. https://doi.org/10.25080/Majora-92bf1922-011

Semenoglou, A.-A., Spiliotis, E., Makridakis, S., & Assimakopoulos, V. (2021). Investigating the accuracy of cross-learning time series forecasting methods. *International Journal of Forecasting*, *37*(3), 1072-1084. https://doi.org/10.1016/j.ijforecast.2020.11.009

Sepúlveda-Rojas, J. P., Rojas, F., Valdés-González, H., & Martín, M. S. (2015). Forecasting Models Selection Mechanism for Supply Chain Demand Estimation. *Procedia Computer Science*, *55*, 1060-1068. https://doi.org/10.1016/j.procs.2015.07.068

Shi, Y., Ding, J., Qu, T., & Xie, B. (2026). Beverage manufacturing demand forecasting system driven by multi-stage machine learning model pool. *Applied Soft Computing*, *187*, 114254. https://doi.org/10.1016/j.asoc.2025.114254

Silver, E. A., Pyke, D. F., & Peterson, R. (1998). *Inventory Management and Production Planning and Scheduling* (Vol. 3). Wiley.

Smola, A. J., & Schölkopf, B. (2004). A tutorial on support vector regression. *Statistics and Computing*, *14*(3), 199-222. https://doi.org/10.1023/B:STCO.0000035301.49549.88

Song, X., Chang, D., Gao, Y., Huang, Q., & Ye, Z. (2025). An aggregate–disaggregate framework for forecasting intermittent demand in fast fashion retailing. *Advanced Engineering Informatics*, *64*, 103069. https://doi.org/10.1016/j.aei.2024.103069

Sousa, M. S., Loureiro, A. L. D., & Miguéis, V. L. (2025). Predicting demand for new products in fashion retailing using censored data. *Expert Systems with Applications*, *259*, 125313. https://doi.org/10.1016/j.eswa.2024.125313

Syntetos, A. A., & Boylan, J. E. (2005). The accuracy of intermittent demand estimates. *International Journal of Forecasting*, *21*(2), 303-314. https://doi.org/10.1016/j.ijforecast.2004.10.001

Syntetos, A. A., Boylan, J. E., & Croston, J. D. (2005). On the categorization of demand patterns. *Journal of the Operational Research Society*, *56*(5), 495-503. https://doi.org/10.1057/palgrave.jors.2601841

Taghiyeh, S., Lengacher, D. C., & Handfield, R. B. (2020). Forecasting model selection using intermediate classification: Application to MonarchFx corporation. *Expert Systems with Applications*, *151*, 113371. https://doi.org/10.1016/j.eswa.2020.113371

Tashman, L. J. (2000). Out-of-sample tests of forecasting accuracy: An analysis and review. *International Journal of Forecasting*, *16*(4), 437-450. https://doi.org/10.1016/S0169-2070(00)00065-0

The pandas development team. (2026). *pandas-dev/pandas: Pandas* (Versión v3.0.3) [Software]. Zenodo. https://doi.org/10.5281/ZENODO.3509134

Theodorou, E., Spiliotis, E., & Assimakopoulos, V. (2025). Forecast accuracy and inventory performance: Insights on their relationship from the M5 competition data. *European Journal of Operational Research*, *322*(2), 414-426. https://doi.org/10.1016/j.ejor.2024.12.033

Tibshirani, R. (1996). Regression Shrinkage and Selection Via the Lasso. *Journal of the Royal Statistical Society Series B: Statistical Methodology*, *58*(1), 267-288. https://doi.org/10.1111/j.2517-6161.1996.tb02080.x

Ulrich, M., Jahnke, H., Langrock, R., Pesch, R., & Senge, R. (2022). Classification-based model selection in retail demand forecasting. *International Journal of Forecasting*, *38*(1), 209-223. https://doi.org/10.1016/j.ijforecast.2021.05.010

Vaswani, A., Shazeer, N., Parmar, N., Uszkoreit, J., Jones, L., Gomez, A. N., Kaiser, Ł. ukasz, & Polosukhin, I. (2017). Attention is All you Need. En I. Guyon, U. V. Luxburg, S. Bengio, H. Wallach, R. Fergus, S. Vishwanathan, & R. Garnett (Eds.), *Advances in Neural Information Processing Systems* (Vol. 30). Curran Associates, Inc. https://proceedings.neurips.cc/paper_files/paper/2017/file/3f5ee243547dee91fbd053c1c4a845aa-Paper.pdf

Virtanen, P., Gommers, R., Oliphant, T. E., Haberland, M., Reddy, T., Cournapeau, D., Burovski, E., Peterson, P., Weckesser, W., Bright, J., Van Der Walt, S. J., Brett, M., Wilson, J., Millman, K. J., Mayorov, N., Nelson, A. R. J., Jones, E., Kern, R., Larson, E., … Vázquez-Baeza, Y. (2020). SciPy 1.0: Fundamental algorithms for scientific computing in Python. *Nature Methods*, *17*(3), 261-272. https://doi.org/10.1038/s41592-019-0686-2

Waskom, M. (2021). seaborn: Statistical data visualization. *Journal of Open Source Software*, *6*(60), 3021. https://doi.org/10.21105/joss.03021

Yasser, H. (2021). *Walmart Sales Prediction—(Best ML Algorithms)* [Dataset]. Kaggle.

Zhao, S., Xie, T., Ai, X., Yang, G., & Zhang, X. (2023). Correcting sample selection bias with model averaging for consumer demand forecasting. *Economic Modelling*, *123*, 106275. https://doi.org/10.1016/j.econmod.2023.106275

Zou, H., & Hastie, T. (2005). Regularization and Variable Selection Via the Elastic Net. *Journal of the Royal Statistical Society Series B: Statistical Methodology*, *67*(2), 301-320. https://doi.org/10.1111/j.1467-9868.2005.00503.x